\documentclass[10pt,twocolumn,letterpaper]{article}

\usepackage[pagenumbers]{cvpr}

\usepackage{graphicx}
\usepackage{amsmath}
\usepackage{amssymb}
\usepackage{booktabs}
\usepackage[table]{xcolor}
\usepackage{adjustbox}
\usepackage[super]{nth}

\usepackage{multirow}

\usepackage{pifont}
\newcommand{\cmark}{\ding{51}}
\newcommand{\xmark}{\ding{55}}
\usepackage{caption}
\usepackage{dblfloatfix}

\newcolumntype{x}[1]{>{\centering\arraybackslash}p{#1pt}}
\newcolumntype{y}[1]{>{\raggedright\arraybackslash}p{#1pt}}
\newcolumntype{z}[1]{>{\raggedleft\arraybackslash}p{#1pt}}

\newlength\savewidth\newcommand\shline{\noalign{\global\savewidth\arrayrulewidth
\global\arrayrulewidth 1pt}\hline\noalign{\global\arrayrulewidth\savewidth}}
\newcommand{\tablestyle}[2]{\setlength{\tabcolsep}{#1}\renewcommand{\arraystretch}{#2}\centering\small}
\makeatletter\renewcommand\paragraph{\@startsection{paragraph}{4}{\z@}
{.4em \@plus1ex \@minus.2ex}{-.5em}{\normalfont\normalsize\bfseries}}\makeatother

\definecolor{citecolor}{RGB}{34,139,34}
\definecolor{citecolor2}{HTML}{0071bc}
\definecolor{lightred}{RGB}{241,140,142}
\definecolor{defaultcolor}{gray}{0.9}
\definecolor{demphcolor}{gray}{.5}

\usepackage[pagebackref=true,breaklinks=true,letterpaper=true,colorlinks,
citecolor=citecolor2,bookmarks=false]{hyperref}
\usepackage[capitalize]{cleveref}
\crefname{section}{\S}{Secs.}
\Crefname{section}{Section}{Sections}
\Crefname{table}{Table}{Tables}
\newcommand{\figref}[1]{Fig.~\ref{#1}}
\newcommand{\secref}[1]{\S\ref{#1}}
\newcommand{\app}{\raise.17ex\hbox{$\scriptstyle\sim$}}
\def\x{$\times$}
\newcommand{\pacc}[1]{{\bf \fontsize{7.5}{42}\selectfont \color{citecolor!80}~(#1)}}
\newcommand{\macc}[1]{{\bf \fontsize{7.5}{42}\selectfont \color{lightred!180}~(#1)}}
\usepackage{textcomp}

\def\cvprPaperID{*****}
\def\confName{****}
\def\confYear{****}

\begin{document}
\title{Masked Feature Prediction for Self-Supervised Visual Pre-Training \vspace{-.8em}}
\author{
Chen Wei\textsuperscript{ *, 1, 2} ~ 
Haoqi Fan\textsuperscript{1} ~ 
Saining Xie\textsuperscript{1} ~ 
Chao-Yuan Wu\textsuperscript{1} ~ 
Alan Yuille\textsuperscript{2} ~  
Christoph Feichtenhofer\textsuperscript{*, 1}\\
\small $^{*}$equal technical contribution \vspace{.5em} \\
\textsuperscript{1}Facebook AI Research \qquad \qquad \textsuperscript{2}Johns Hopkins University
\vspace{-1em}
}
\maketitle

\begin{abstract}
We present Masked Feature Prediction (MaskFeat) for self-supervised pre-training of video models. Our approach first randomly masks out a portion of the input sequence and then predicts the \emph{feature} of the masked regions. We study five different types of features and find Histograms of Oriented Gradients (HOG), a hand-crafted feature descriptor, works particularly well in terms of both performance and efficiency. 
We observe that the local contrast normalization in HOG is essential for good results, which is in line with earlier work using HOG for visual recognition. 
Our approach can learn abundant visual
knowledge and drive large-scale Transformer-based models. Without using extra model weights or supervision, \mbox{MaskFeat} pre-trained on unlabeled videos achieves unprecedented results of 86.7\% with \mbox{MViT-L} on Kinetics-400, 88.3\% on Kinetics-600, 80.4\% on Kinetics-700, 39.8 mAP on AVA, and 75.0\% on SSv2. 
MaskFeat further generalizes to image input, which can be interpreted as a video with a single frame and obtains competitive results on ImageNet.
\end{abstract}

\section{Introduction}
\label{sec:intro}

\begin{figure}[t]
\centering
\subfloat{\begin{tabular}{ccc}\quad\small{masked input} & \quad\small{HOG prediction} & \quad\small{original image}\end{tabular}}

\subfloat{\includegraphics[width=0.98\linewidth]{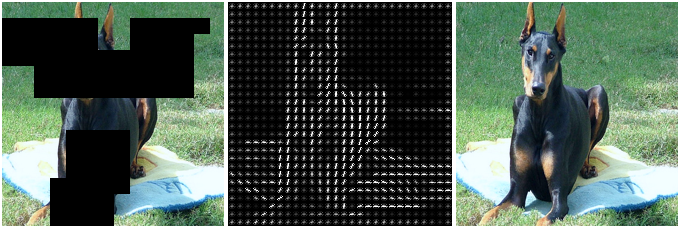}}

\subfloat{\includegraphics[width=0.98\linewidth]{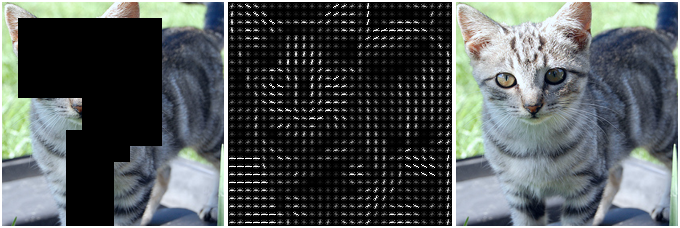}}

\subfloat{\includegraphics[width=0.98\linewidth]{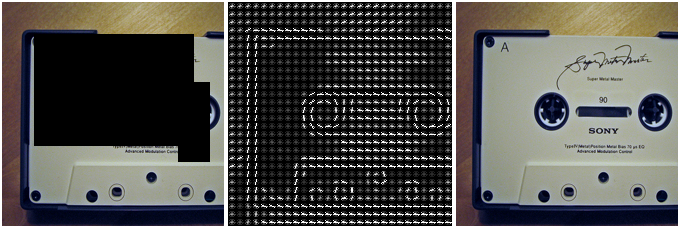}}

\vspace{-7pt}
\caption{\textbf{Example HOG predictions} on unseen \textit{validation} input. Our model is learned by predicting  features (\textit{middle}) given masked inputs (\textit{left}). Original images (\textit{right}) are not used for prediction. More qualitative examples for video and image are in \cref{fig:more-viz-image,fig:more-viz-video}.}
\label{fig:vis}
\vspace{-13pt}
\end{figure}

Self-supervised pre-training has been phenomenally successful in natural language processing powering large-scale Transformers~\cite{transformer} with billion-scale data~\cite{bert, gpt-3}. The underlying idea is an astonishingly simple \textit{mask-and-predict} task, that is, first masking out some tokens within a text and then predicting the invisible content given the visible text. 

Humans have a remarkable ability
to predict how the world appears and moves when observing it as a continuous stream of spatiotemporal information. 
Consider the examples in the \nth{1} column of \cref{fig:vis}. Even without seeing the masked content, we are able to understand the object structure and draw a rough outline or silhouette of imagined information (up to some details), by using visual knowledge about the visible structures. In this work, we show that predicting certain masked features (\eg gradient histograms in the \nth{2} column) can be a powerful objective for self-supervised visual pre-training, especially in the video domain which contains rich visual information.

One essential difference between vision and language is that vision has no pre-existing vocabulary to shape the prediction task into a well-defined classification problem. In contrast, the raw spatiotemporal visual signal is continuous and dense posing a major challenge to masked visual prediction.
One immediate solution is to imitate the language vocabulary by building a visual vocabulary that discretizes frame
patches into tokens, as explored in BEiT~\cite{dall-e, beit}.
However, this requires an external tokenizer which can be limited in compute-intensive video understanding scenario.

We present \textbf{Mask}ed \textbf{Feat}ure Prediction (MaskFeat), a pre-training objective that directly regresses \textit{features} of the masked content. Specifically, our approach ingests the masked space-time input with a vision Transformer backbone~\cite{vit,mvitv2} and predicts a certain feature representation of the masked content. In this way, the pre-trained model acquires an adequate understanding of the complex space-time structures within dense visual signals.

We study a broad spectrum of feature types, from pixel colors and hand-crafted feature descriptors, to discrete visual tokens, activations of deep networks, and pseudo-labels from network predictions. Our study reveals:

(\textit{i}) Simple histogram of oriented gradients (center column in \cref{fig:vis}), as in the popular HOG~\cite{hog} and SIFT~\cite{sift} descriptors which dominated visual recognition for over a decade, is a particularly effective target for MaskFeat in terms of both performance and efficiency.

(\textit{ii}) The discretization (tokenization) of visual signals is not necessary for masked visual prediction, and continuous \textit{feature regression} (\ie MaskFeat) can work well.

(\textit{iii}) Semantic knowledge from human annotations is not always helpful for MaskFeat, but characterizing local patterns seems important. For example, predicting supervised features from CNNs or ViTs trained on \textit{labeled} data leads to \textit{degraded} performance. 

Our approach is conceptually and practically simple. Compared to contrastive methods that require a siamese structure and two or more views of each training sample (\eg, \cite{moco,simsiam,videomoco}), MaskFeat uses a \textit{single network} with a \textit{single view} of each sample; and unlike contrastive methods that strongly rely on carefully designed data augmentation, MaskFeat works fairly well with minimal augmentation. 

Compared to previous masked visual prediction methods~\cite{beit,vimpac}, MaskFeat with HOG does \textit{not involve any external model}, such as a dVAE tokenizer~\cite{dall-e} that introduces not only an extra pre-training stage on 250M images, but also non-negligible training overhead in masked modeling. 

We show that MaskFeat can pre-train large-scale video models that generalize well.
Transformer-based video models, though powerful, are previously known to be prone to over-fitting and heavily rely on \textit{supervised} pre-training~\cite{mvitv2,vivit} on large-scale \textit{image} datasets, \eg, ImageNet-21K (IN-21K)~\cite{imagenet}. While MaskFeat opens the door for directly pre-training on unlabeled videos which shows enormous benefits for video understanding.

Our results on standard video benchmarks are groundbreaking: MaskFeat pre-trained \mbox{MViT-L}~\cite{mvitv2} gets \textbf{86.7}\% top-1 accuracy on Kinetics-400~\cite{k400} without using any external data, greatly surpassing the best prior number of this kind by +\textbf{5.2}\%, and also methods using large-scale image datasets, \eg, IN-21K and JFT-300M~\cite{jft}. When transferring to downstream tasks, MaskFeat gets unprecedented results of \textbf{39.8} mAP on action detection (AVA~\cite{ava}) and \textbf{75.0}\%  top-1 accuracy on human-object interaction classification (SSv2~\cite{ssv2}). When generalized to the image domain, MaskFeat also obtains competitive 84.0\% top-1 with ViT-B and 85.7\% with ViT-L using only ImageNet-1K~\cite{imagenet}.

Our code will be available in PyTorchVideo\footnote{{\scriptsize\url{https://github.com/facebookresearch/pytorchvideo}}}$^\text{,}$\footnote{\scriptsize\url{https://github.com/facebookresearch/SlowFast}}~\cite{fan2021pytorchvideo,fan2020pyslowfast}.

\begin{figure}[t!]
\centering
\includegraphics[width=0.9\linewidth]{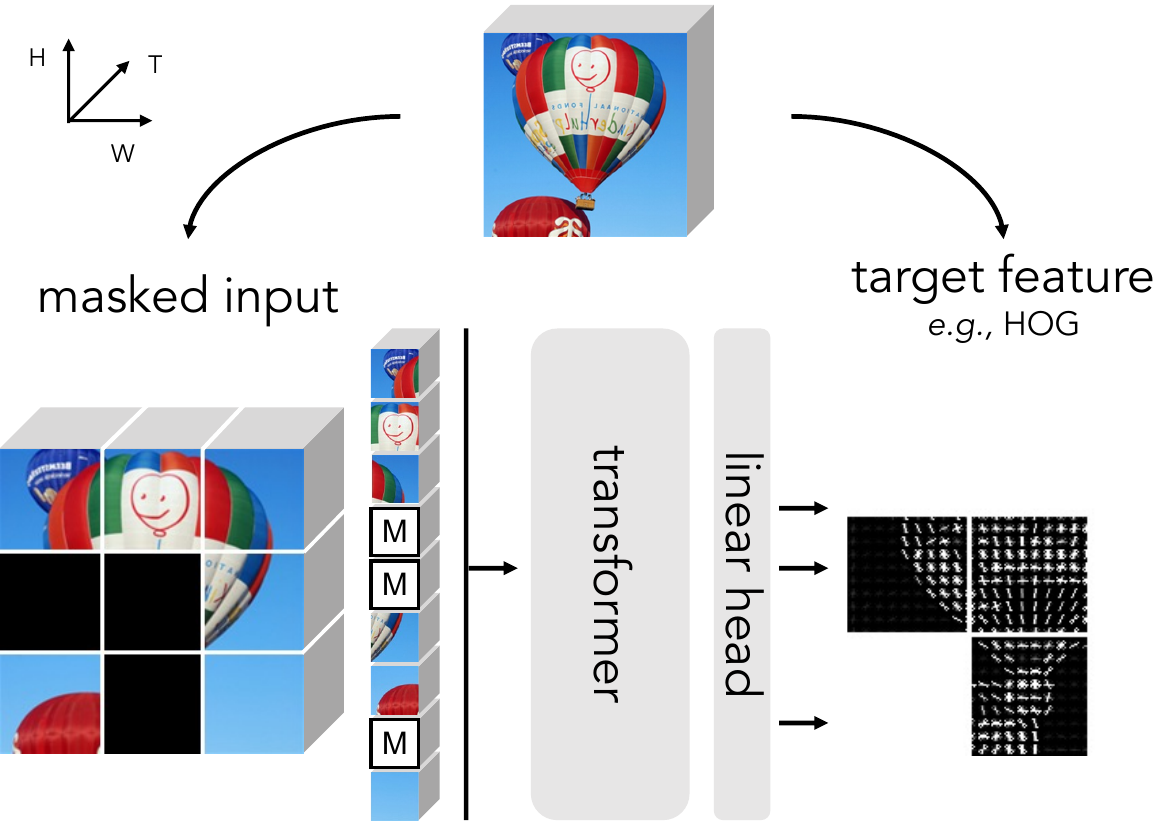}
\vspace{-7pt}
\caption{\textbf{MaskFeat pre-training.} We randomly replace the input space-time cubes of a video with a $\texttt{[MASK]}$ token and directly regress features (\eg HOG) of the masked regions. After pre-training, the  Transformer is fine-tuned on end tasks.
\label{fig:framework}}
\vspace{-10pt}
\end{figure}

\section{Method}
\label{sec:method}

We start by describing MaskFeat and its instantiations for video and image understanding in \secref{subsec:mfp}. We then introduce and discuss five candidates for target features in \secref{subsec:target-feat}.

\subsection{Masked Feature Prediction}
\label{subsec:mfp}

Our method performs a masked visual prediction task, motivated by humans' ability to inpaint masked visual content up to some details. The task first randomly masks out a few space-time cubes of a video, and then predicts the masked ones given the remaining ones. By modeling masked samples, the model attains video understanding in the sense of recognizing parts and motion of objects.
For instance, to solve the examples in \cref{fig:vis,fig:more-viz-video}, a model has to first recognize the objects based on the visible area, and also know what the objects typically \textit{appear} and how they usually \textit{move} to inpaint the missing area.

One key component of the task is the prediction target. Masked language modeling tokenizes the corpus with a vocabulary to serve as the target~\cite{bert}. In contrast, the raw visual signal is continuous and high-dimensional and there is no natural vocabulary available. In MaskFeat, we propose to predict \textit{features} of the masked area. And the supervision is provided by features extracted from the original, intact sample. We use a wide interpretation of features~\cite{chatfield2014return}, from hand-crafted feature descriptors, to activations of deep networks. The choice of the target feature largely defines the task and impacts the property of the pre-trained model, which we discuss in \secref{subsec:target-feat}.

\paragraph{Instantiations.}

We first describe MaskFeat for video input.

A video is first divided into space-time cubes as in typical video Vision Transformers~\cite{mvit,mvitv2}. The cubes are then projected (\ie convolved) to a sequence of tokens. To perform masking, some of the tokens in the sequence are randomly masked out by being replaced with a $\texttt{[MASK]}$ token. This is a learnable embedding indicating masked patches. A block of tokens is masked together which we detail in \secref{sec:masking-strategy}.
To make a prediction, the token sequence after $\texttt{[MASK]}$ token replacement, with positional embedding added, is processed by the Transformer. Output tokens corresponding to the masked cubes are projected to the prediction by a linear layer. The prediction is simply the feature of the 2-D spatial patch temporally centered in each masked cube (see discussions in \cref{app:video-target}). The number of output channels is adjusted to the specific target feature (\eg, 3\x16\x16 if predicting RGB colors of pixels in a 16\x16 patch). The loss is only operated on the masked cubes. Our instantiation is inspired by BERT~\cite{bert} and BEiT~\cite{beit}, illustrated in \figref{fig:framework}.

MaskFeat can be easily instantiated in the image domain, which can be interpreted as a video with one single frame. Most operations are shared, except that there is no temporal dimension and each token now represents only a spatial patch instead of a space-time cube.

\subsection{Target Features}
\label{subsec:target-feat}

We consider five different types of target features. The targets are categorized into two groups: 1) one-stage targets that can be directly obtained including pixel colors and HOG, and 2) other two-stage targets extracted by a trained deep network or \textit{teacher}. As predicting two-stage targets is effectively learning from a trained deep network teacher, it resembles a form of model distillation~\cite{distillation}; thereby, an extra computational cost of pre-training and inference of the teacher model is inevitable. The five feature types are:

\paragraph{Pixel colors.}
The most straightforward target is arguably the colors of video pixels. Specifically, we use RGB values that are normalized by the mean and the standard deviation of the dataset. We minimize the $\ell_2$ distance between the model's prediction and the ground-truth RGB values. A similar idea has been explored in \cite{inpainting} as a image inpainting task and in~\cite{beit,vit} for masked image prediction. Though simple, pixels as target have a potential downside of overfitting to local statistics (\eg illumination and contrast variations) and high-frequency details, which are presumably insignificant~\cite{improved-gan} for interpretation of visual content.

\paragraph{HOG.} 
Histograms of Oriented Gradients (HOG)~\cite{hog} is a feature descriptor that describes the distribution of gradient orientations or edge directions within a local subregion. A HOG descriptor is implemented by a simple gradient filtering (\ie subtracting neighboring pixels) to compute magnitudes and orientations of gradients at each pixel. The gradients within each small local subregion or \textit{cell} are then accumulated into orientation histogram vectors of several bins, voted by gradient magnitudes. The histogram is normalized to unit length. These features are also used in well-known SIFT~\cite{sift} descriptors for detected keypoints or in a dense fashion for classification~\cite{chatfield2014return}. Similarly, we extract HOG on a dense grid for the whole image, which suits the prediction target for randomly masked patches.

HOG is characteristic of capturing local shapes and appearances while being partially invariant to geometric changes as long as translations are within the spatial cell and rotations are smaller than orientation bin size. Further, it provides invariance to photometric changes as image gradients and local contrast normalization absorb brightness (\eg illumination) and foreground-background contrast variation. These invariances are vital for good results when using HOG for pedestrian detection in both image~\cite{hog} and video~\cite{hogvideo} domains. In accordance to this, our studies (\secref{sec:ablations}) reveal local-contrast normalization in HOG is also essential for MaskFeat pre-training.

Finally, HOG computation is cheap and introduces \textit{negligible} overhead. It can be implemented as a two-channel convolution to generate gradients in \textit{x} and \textit{y} axis (or by subtracting neighboring horizontal and vertical pixels)
, followed by histogramming and normalization.

Our method then simply predicts the histograms summarizing masked patches. Instead of computing HOG only on masked patches, we first obtain a HOG feature map on the whole image and then split the map into patches. In this way, we reduce padding on boundaries of each masked patch. The histograms of masked patches are then flattened and concatenated into a 1-D vector as the target feature. Our loss minimizes the $\ell_2$ distance between the predicted and original HOG feature. We collect HOG in each RGB channel to include color information which can slightly improve its performance (\secref{sec:ablations}).  

\begin{table*}[!b]
\centering
\vspace{-10pt}
\renewcommand\thetable{2}
\tablestyle{4.5pt}{1.05}
\subfloat{\includegraphics[width=0.27\textwidth]{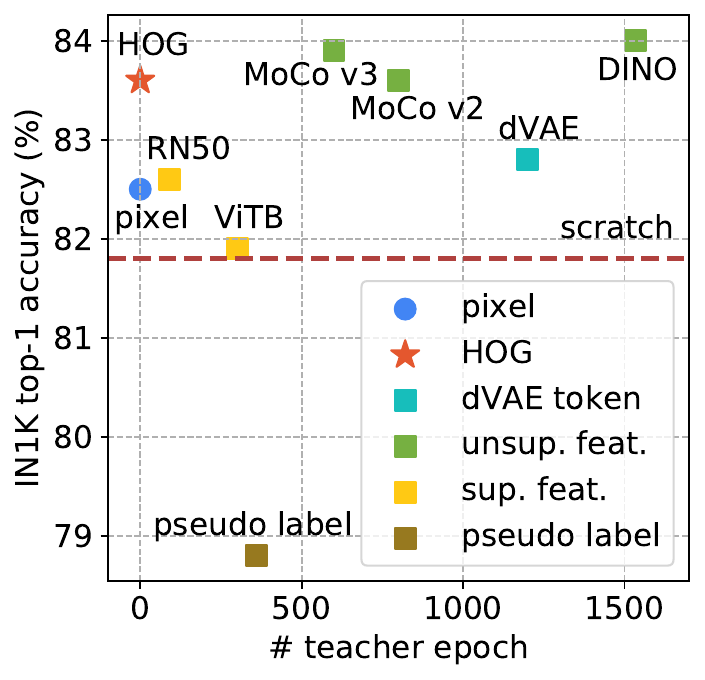}}
\subfloat{
\begin{tabular}{lc|lccc|c}
feature type     & one-stage   & variant                             & arch.    & param. &  epoch$^\dagger$ & top-1 \\
\shline
scratch              & -           & DeiT~\cite{deit}                    & -        & -      & -                & 81.8 \\
\hline
pixel colors         & \cmark      & RGB                                 & -        & -      & -                & 82.5 \\
\rowcolor{defaultcolor}
image descriptor     & \cmark      & HOG~\cite{hog}                      & -        & -      & -                & \textbf{83.6} \\
\hline
dVAE token           & \xmark      & DALL-E~\cite{dall-e}                & dVAE     & 54     & 1199             & 82.8 \\
unsupervised feature & \xmark      & MoCo v2~\cite{mocov2}               & ResNet50 & 23     & 800              & 83.6 \\
unsupervised feature & \xmark      & MoCo v3~\cite{mocov3}               & ViT-B    & 85     & 600              & 83.9 \\
unsupervised feature & \xmark      & DINO~\cite{dino}                    & ViT-B    & 85     & 1535             & \textbf{84.0} \\
supervised  feature  & \xmark      & pytorch~\cite{pytorch}              & ResNet50 & 23     & 90               & 82.6 \\
supervised  feature  & \xmark      & DeiT~\cite{deit}                    & ViT-B    & 85     & 300              & 81.9 \\
pseudo-label         & \xmark      & Token Labeling~\cite{tokenlabeling} & NFNet-F6 & 438    & 360              & 78.8 \\
\end{tabular}}
\vspace{-10pt}
\caption{\textbf{Comparing target features for MaskFeat (\textit{image}).} For all targets, ViT-B is pre-trained with MaskFeat for 300 epochs on IN-1K. We report 100-epoch fine-tuning accuracy on IN-1K. For two-stage targets, we report the \textit{teacher} architecture, number of parameters (M), and effective epoch$^\dagger$ on IN-1K. The default entry is marked in \colorbox{defaultcolor}{gray}. The plot on the left visualizes the acc/epoch trade-off of the table.}
\raggedright
\footnotesize{$^\dagger$ Different teachers use different training strategies. dVAE is pre-trained on an external 250M dataset, while self-supervised methods require multi-view training. To measure the cost in a unified way, we normalize the number of epochs by the cost of one epoch on IN-1K training set with \textit{one 224$^2$ view}.}
\label{tab:feat-img}
\vspace{-14pt}
\end{table*}

\paragraph{Discrete variational autoencoder (dVAE).}
To address the continuous high-dimensional nature of visual signals, DALL-E~\cite{dall-e} proposes to compress an image with a dVAE codebook. In particular, each patch is encoded into a token which can assume 8192 possible values using a pre-trained dVAE model. Now the task is to predict the categorical distribution of the masked token by optimizing a cross-entropy loss, as explored in BEiT~\cite{beit}. However, there is an extra computational cost induced by pre-training the dVAE and tokenizing images alongside masked feature prediction.

\paragraph{Deep features.}
In comparison to discretized tokens, we consider directly using continuous deep network features as the prediction target.  We use a pre-trained model to produce features as a teacher, either a CNN or ViT, and our loss minimizes the cosine distance (\ie mean squared error of $\ell_2$-normalized features).

For CNN teachers, we use the last layers' features corresponding to the masked patches and for ViT we use the respective output patch tokens. We mainly compare features from self-supervised models, which are considered to contain more diverse scene layout~\cite{dino} and preserve more visual details~\cite{good-transfer} than features from supervised models. (Though, the usage of human annotations makes the pre-training technically not self-supervised.) Supervised features are expected to be more semantic as they are trained through human annotations. Similar to dVAE, a non-trivial amount of extra computation is involved when using extra model weights for masked feature generation.

\paragraph{Pseudo-label.}
To explore an even more high-level semantic prediction target, we consider predicting class labels of masked patches. We utilize labels provided by Token Labeling~\cite{tokenlabeling}, where each patch is assigned an individual location-specific IN-1K pseudo-label. This class label map is generated by a pre-trained high-performance supervised deep network~\cite{nfnet} teacher. The masked feature prediction stage is optimized by a cross-entropy loss.

We next study the features discussed in this section.

\section{Study: Target Features for MaskFeat}
\label{sec:study}

\begin{table}[b!]
\centering
\vspace{-15pt}
\renewcommand\thetable{1}
\tablestyle{5.5pt}{1.05}
\begin{tabular}{lc|l|c}
\small{feature type} & \small{one-stage} & \small{variant}         &  \small{top-1} \\
\shline
scratch                  & -                 & MViT-S~\cite{mvitv2}    &  81.1 \\
\hline
pixel                    & \cmark            & RGB                     &  80.7 \\
\rowcolor{defaultcolor}
image descriptor         & \cmark            & HOG~\cite{hog}          &  \textbf{82.2}  \\
\hline
dVAE                     & \xmark            & DALL-E~\cite{dall-e}    &  81.7 \\
unsupervised feature     & \xmark            & DINO~\cite{dino}, ViT-B &  \textbf{82.5} \\
supervised feature       & \xmark            & MViT-B~\cite{mvit}      &  81.9     \\
\end{tabular}
\vspace{-10pt}
\caption{\textbf{Comparing target features for MaskFeat (\textit{video}).} All variants are pre-trained with MaskFeat for 300 epochs on MViT-S, 16\x4. We report fine-tuning accuracy on K400. Default is \colorbox{defaultcolor}{gray}.}
\label{tab:feat-vid}
\vspace{-5pt}
\end{table}

\paragraph{Settings.}

We use a pre-training and fine-tuning protocol, following BEiT~\cite{beit}. We pre-train \mbox{MViT-S, 16\x4}~\cite{mvitv2} with MaskFeat on Kinetics-400 (K400)~\cite{k400} training set for 300 epochs. We also apply MaskFeat on images, where we pre-train ViT-B~\cite{vit} on the ImageNet-1K (IN-1K)~\cite{imagenet} training set for 300 epochs. We report top-1 fine-tuning accuracy (\%) on both datasets. We pre-train and fine-tune all targets with the same recipe which we find generally good in practice (\S\ref{app:detail-cls}). For targets that involve a teacher model, we use official models released by the authors.

Most features are compared on both video and image domains except pseudo-label for which the pseudo-label map is only available on IN-1K~\cite{tokenlabeling}. Results are summarized in Tables \ref{tab:feat-vid} (video) and \ref{tab:feat-img} (image), analyzed next: 

\paragraph{One-stage methods.}
The fine-tuning accuracy for pixel color prediction in Tables ~\ref{tab:feat-vid} \& \ref{tab:feat-img} shows, that compared to the from-scratch baselines, regressing RGB colors produces a slight drop of -0.4\% for video classification and a relatively small gain of +0.7\% for image. Even though our predicting pixel colors result on IN-1K (82.5\%) is better than that reported in BEiT~\cite{beit} (81.0\%), we similarly observe that pixel values are not ideal direct targets, presumably because they are considered to be too explicit~\cite{dall-e}.
In comparison, HOG, by summarizing the local gradient distribution, contributes to large improvements of +\textbf{1.1}\% on K400 and +\textbf{1.8}\% on IN-1K over the from-scratch baselines without any extra model which is typical in two-stage methods.

\paragraph{Two-stage methods.}
First, dVAE improves by +0.6\% for K400 and +1.0\% for IN-1K over their from-scratch baselines. This is better than pixel colors, but outperformed by HOG which does not use an external model.

Next, compared to dVAE, we study MaskFeat to predict \textit{continuous}, unsupervised features: We compare DINO~\cite{dino} (with ViT-B) and MoCo~\cite{mocov2, mocov3} (with ResNet50~\cite{resnet} and ViT-B), all pre-trained on IN-1K, even for the video pre-training. Unsupervised features contribute a notable gain for both video and image classification: The DINO variant achieves a gain of +1.4\% on K400 and +2.2\% on IN-1K compared to their baselines. However, this approach has two main drawbacks, (i) the unsupervised feature extractor needs to be pre-trained \eg worth over thousand epochs in the case of DINO, (ii) the unsupervised features need to be computed on the target data. Still, MaskFeat w/ DINO and MoCo v3 features boosts their original accuracy~\cite{dino,mocov3}.

Finally, supervised features (from ResNet50 or ViT-B) as well as token labels, though utilizing human annotations, lag behind unsupervised features and HOG. In fact, we notice \textit{significant over-fitting} during fine-tuning for supervised features and token labels, suggesting that predicting features learned from class labels is not suitable in MaskFeat. We hypothesize that class label being invariant to local shapes and textures of the same object disables the ability of MaskFeat to model object's internal structure.

\paragraph{Discussion.}
Our results suggest that a broad spectrum of image features can serve as targets in masked visual prediction, and provide gains over the train-from-scratch baseline. We find that although masked language modeling~\cite{bert} originally predicts the categorical distribution over a pre-defined vocabulary, discretization as in BEiT~\cite{beit} is not required for vision. We find that continuous unsupervised features and image descriptors can be strong prediction targets, while the latter come without cost compared to the former which also entail a form of \textit{model distillation}~\cite{hinton2015distilling,deit}. An interesting observation is that \textit{supervisedly} trained target features produce poor results, which might relate to class-level specific information being present in features~\cite{zhou2014object,Bau_2017_CVPR} that is too global for local mask modeling. 
Overall, considering the trade-off between performance and computational cost, predicting HOG holds a good balance and therefore we use it as default feature for MaskFeat in the following sections.

\section{Experiments: Video Recognition}

\paragraph{Settings.}
We evaluate with base and large models of improved MViT~\cite{mvitv2}. The original MViT in \cite{mvit} is termed as MViTv1. The models are pre-trained \textit{only} on video clips in the training set of K400~\cite{imagenet} without labels. Our augmentation includes random resized cropping and horizontal flipping. Our models are pre-trained and fine-tuned at 224$^2$ resolution if not specified. We randomly mask out 40\% of total space-time cubes with \textit{cube} masking detailed in \secref{sec:masking-strategy}. More implementation details are in \cref{app:detail-cls}.

\subsection{Main Results on Kinetics}

\paragraph{Kinetics-400.}
\cref{tab:k400-finetune} compares MaskFeat with prior work on K400 dataset. From top to bottom, it has three sections.

The first section presents prior work using CNNs, which commonly do not use any pre-training. The second section presents representative Transformer-based methods, most of which are heavily dependent on supervised pre-training on \textit{large-scale image datasets}.

The third section provides direct comparisons on MViT models. Note that these models are strong baselines and are state-of-the-art for training-from-scratch on their own. Still, 300 epochs of MaskFeat pre-training improve the scratch \mbox{MViT-S, 16\x4}~\cite{mvitv2} with 81.1\% top-1 accuracy by +1.1\%. Here, the suffix 16\x4 represents that the model takes 16 frames with a temporal stride of 4 as input during training.

Next, we explore larger models for which supervised IN-21K pre-training is popular. Pre-trained with MaskFeat for 800 epochs on K400, the large model \mbox{MViT-L, 16\x4} reaches 84.3\% top-1, outperforming its scratch baseline by a large margin of +\textbf{3.8}\% and its IN-21K supervised counterpart by +0.8\%. Similar to the image domain, MaskFeat is more significant with larger models, showing that our approach is salable to model capacity. The result also suggests that MaskFeat adapts to different \textit{model types}, as MViT is a Transformer model \textit{with convolutions}. We provide ablation on the pre-training schedule in \cref{sec:video-schedule}.

We further explore the data scalability of MaskFeat. In particular, we pre-train \mbox{MViT-L, 16\x4} with Kinetics-600 (K600)~\cite{k600} containing \raisebox{0.5ex}{\texttildelow}387K training videos, 1.6\x~more than K400. We pre-train for 300 epochs on K600 to use a slightly smaller training budget as the 800 epochs on K400. We again fine-tune on K400 and observe that pre-training on K600, without any labels, contributes to another +0.8\% gain over K400 pre-training to reach 85.1\% top-1.

\begin{table}[t!]
\vspace{-10pt}
\centering
\captionsetup[sub]{font=normalsize}
\tablestyle{2.0pt}{1.02}
\resizebox{1.02\linewidth}{!}{
\hspace*{-7pt}
\begin{tabular}{l|l|c|c|r|r}
model & pre-train & top-1 & top-5 & \footnotesize{FLOPs}\x views & \footnotesize{Param} \\
\shline \hline
Two-Stream I3D~\cite{two-stream} &\multicolumn{1}{c|}{-} & 71.6 & 90.0 & 216~\x~NA & 25 \\
ip-CSN-152~\cite{ipcsn} &\multicolumn{1}{c|}{-} & 77.8 & 92.8 & 109\x3\x10 & 33 \\
{SlowFast} {\scriptsize{16\x 8 +NL}}~\cite{slowfast} &\multicolumn{1}{c|}{-} & 79.8 & 93.9 & 234\x3\x10 & 60 \\
X3D-XL~\cite{x3d} &\multicolumn{1}{c|}{-} & 79.1 & 93.9 & 48\x3\x10 & 11 \\ 
MoViNet-A6~\cite{movinet} & \multicolumn{1}{c|}{-} & 81.5 & 95.3 & 386\x1\x1 & 31 \\
\shline\hline
MViTv1-B, 64\x3~\cite{mvit} &\multicolumn{1}{c|}{-} & 81.2 & 95.1 & 455\x3\x3 & 37 \\
Swin-B, 32\x2~\cite{video-swin} & Sup., IN-21K & \color{demphcolor}{82.7} & \color{demphcolor}{95.5} & 282\x3\x4 & 88 \\
\hline
ViT-B-TimeSformer~\cite{timesformer} & Sup., IN-21K & \color{demphcolor}{80.7} & \color{demphcolor}{94.7} & 2380\x3\x1 & 121 \\
Swin-L, 32\x2~\cite{video-swin} & Sup., IN-21K & \color{demphcolor}{83.1} & \color{demphcolor}{95.9} & 604\x3\x4 & 197 \\
ViViT-L~\cite{vivit} & Sup., JFT-300M & \color{demphcolor}{83.5} & \color{demphcolor}{94.3} & 3980\x3\x1 & 308 \\
\hline
Swin-L\textuparrow384, 32\x2~\cite{video-swin} & Sup., IN-21K & \color{demphcolor}{84.9} & \color{demphcolor}{96.7} & 2107\x5\x10& 200 \\
ViViT-H~\cite{vivit} & Sup., JFT-300M & \color{demphcolor}{84.9} & \color{demphcolor}{95.8} & 3981\x3\x4 & 654 \\
TokenLearner~\cite{tokenlearner} & Sup., JFT-300M & \color{demphcolor}{85.4} & \color{demphcolor}{N/A} & 4076\x3\x4 & 450 \\
Florence\textuparrow384~\cite{florence} & {Text, FLD-900M} & \color{demphcolor}{86.5} & \color{demphcolor}{97.3} & N/A\x3\x4 & 647\\
\hline
\multirow{2}{*}{SwinV2-G\textuparrow384~\cite{swinv2}} & {MIM + Sup.} & \multirow{2}{*}{\color{demphcolor}{86.8}} & \multirow{2}{*}{\color{demphcolor}{N/A}} & \multirow{2}{*}{N/A\x5\x4} & \multirow{2}{*}{3000} \\
& {IN-21K+Ext-70M} & & & \\
\shline\hline
MViT-S, 16\x4~\cite{mvitv2} &\multicolumn{1}{c|}{-} & 81.1 & 94.9 & 71\x1\x10 & 36 \\
MViT-S, 16\x4~\cite{mvitv2} & Sup., IN-21K & \color{demphcolor}{82.6} & \color{demphcolor}{95.3} & 71\x1\x10 & 36 \\
MViT-S, 16\x4~\cite{mvitv2} & \textbf{MaskFeat}, K400 & \textbf{82.2} & \textbf{95.1} & 71\x1\x10 & 36 \\
\hline
MViT-L, 16\x4~\cite{mvitv2} &\multicolumn{1}{c|}{-} & 80.5 & 94.1 & 377\x1\x10 & 218 \\
MViT-L, 16\x4~\cite{mvitv2} & Sup., IN-21K & \color{demphcolor}{83.5} & \color{demphcolor}{95.9} & 377\x1\x10 & 218 \\
MViT-L, 16\x4~\cite{mvitv2} & \textbf{MaskFeat}, K400 & \textbf{84.3} & \textbf{96.3} & 377\x1\x10 & 218 \\
MViT-L, 16\x4~\cite{mvitv2} & \textbf{MaskFeat}, K600 & \color{demphcolor}{85.1} & \color{demphcolor}{96.6} & 377\x1\x10 & 218 \\
\hline
MViT-L\textuparrow312, 32\x3~\cite{mvitv2} & \multicolumn{1}{c|}{-} & 82.2 & 94.7 & 2063\x3\x5 & 218 \\
MViT-L\textuparrow312, 32\x3~\cite{mvitv2} & Sup., IN-21K & \color{demphcolor}{85.3} & \color{demphcolor}{96.6} & 2063\x3\x5 & 218 \\
MViT-L\textuparrow312, 32\x3~\cite{mvitv2} & \textbf{MaskFeat}, K400 & \textbf{86.3} & \textbf{97.1} & 2063\x3\x5 & 218 \\
\hline
MViT-L\textuparrow312, 40\x3~\cite{mvitv2} & \textbf{MaskFeat}, K400 & 86.4 & 97.1 & 2828\x3\x4 & 218 \\
MViT-L\textuparrow352, 40\x3~\cite{mvitv2} & \textbf{MaskFeat}, K400 & \textbf{86.7} & \textbf{97.3} & 3790\x3\x4 & 218 \\
MViT-L\textuparrow352, 40\x3~\cite{mvitv2} & \textbf{MaskFeat}, K600 & \color{demphcolor}{\textbf{87.0}} & \color{demphcolor}{\textbf{97.4}} & 3790\x3\x4 & 218 \\
\end{tabular}
}

\vspace{-10pt}
\caption{\textbf{Comparison with previous work on Kinetics-400}. We report the inference cost with a single ``view" (temporal clip with spatial crop) \x { }the number of views (FLOPs\x view$_\text{space}$\x view$_\text{time}$).
Each ``view'' consists of $T$ frames with $\tau$ temporal stride, $T\times\tau$. Magnitudes are Giga ($10^9$) for FLOPs and Mega ($10^6$) for Param. Accuracy of models trained with external data is \color{demphcolor}{de-emphasized}.}
\label{tab:k400-finetune}
\vspace{-10pt}
\end{table}

Next, we fine-tune the 84.3\% top-1 \mbox{MViT-L, 16\x4} MaskFeat model for 30 epochs to larger spatial sizes of 312$^2$ and 352$^2$, as well as longer temporal durations of 32 and 40 frames with a temporal stride of three. The resulting extra large model \mbox{MViT-L\textuparrow352, 40\x3}, \textit{without using any external data}, achieves a top accuracy of \textbf{86.7}\%. Previously, Transformer-based video models heavily rely on supervised pre-training on large \textit{image} datasets to reach high accuracy. For example, 84.9\% top-1 \mbox{Swin-L\textuparrow384}~\cite{video-swin} with IN-21K and 84.9\% ViViT-H~\cite{vivit} with JFT-300M~\cite{jft}.
MaskFeat opens the door for directly pre-training on unlabeled videos which shows enormous benefits for video understanding, as we can boost the previous best accuracy without external data on K400 (81.5\% \mbox{MoViNet-A6}~\cite{movinet}) by +5.2\%.

Our best \textbf{87.0}\% top-1 accuracy is achieved by fine-tuning the 85.1\% \mbox{MViT-L, 16\x4} pre-trained with MaskFeat on 387K training videos in K600 \textit{using no labels}.

Our results with just K400 (86.7\%) is already similar to recent 86.5\% Florence~\cite{florence} and 86.8\% SwinV2-G~\cite{swinv2}. Florence uses 900M curated text-image pairs. SwinV2-G utilizes a giant model with three billion parameters, and is first self-supervisedly then supervisedly pre-trained on a large dataset of IN-21K plus 70M in-house images. The efficiency of our approach in terms of parameter count, compute cost, data, and annotation suggests again the advantage of MaskFeat \textit{directly} pre-training on \textit{unlabeled videos}.

\begin{table}[t!]
\centering
\captionsetup[sub]{font=normalsize}
\tablestyle{2.0pt}{1.02}
\resizebox{1.02\linewidth}{!}{
\subfloat[\textbf{Kinetics-600}\label{tab:k600-finetune}]{
\hspace*{-7pt}
\begin{tabular}{l|l|c|c|r|r}
model & pre-train & top-1 & top-5 & \footnotesize{FLOPs}\x views & \footnotesize{Param} \\
\shline \hline
{SlowFast} {\scriptsize{16\x 8 +NL}}~\cite{slowfast} &\multicolumn{1}{c|}{-} & 81.8 & 95.1 & 234\x3\x10 & 60 \\
X3D-XL~\cite{x3d} &\multicolumn{1}{c|}{-} & 81.9 & 95.5 & 48\x3\x10 & 11 \\ 
MoViNet-A6~\cite{movinet} & \multicolumn{1}{c|}{-} & 84.8 & 96.5 & 386\x1\x1 & 31 \\
\shline\hline
MViTv1{\scriptsize-B-24}, 32\x3~\cite{mvit} &\multicolumn{1}{c|}{-} & 84.1 & 96.5 & 236\x1\x5 & 53 \\
Swin-B, 16\x2~\cite{video-swin} & Sup., IN-21K & \color{demphcolor}{84.0} & \color{demphcolor}{96.5} & 282\x3\x4 & 88 \\
Swin-L\textuparrow384, 32\x2~\cite{video-swin} & Sup., IN-21K & \color{demphcolor}{86.1} & \color{demphcolor}{97.3} & 2107\x5\x10& 200 \\
ViViT-H~\cite{vivit} & Sup., JFT-300M & \color{demphcolor}{85.8} & \color{demphcolor}{96.5} & 3981\x3\x4 & 654 \\
Florence\textuparrow384~\cite{florence} & {Text, FLD-900M} & \color{demphcolor}{87.8} & \color{demphcolor}{97.8} & N/A\x3\x4 & 647\\
\shline\hline
MViT-L, 16\x4~\cite{mvitv2} & Sup., IN-21K & \color{demphcolor}{85.8} & \color{demphcolor}{97.1} & 377\x1\x10 & 218 \\
MViT-L, 16\x4~\cite{mvitv2} & \textbf{MaskFeat}, K600 & \textbf{86.4} & \textbf{97.4} & 377\x1\x10 & 218 \\
\hline
MViT-L\textuparrow312, 40\x3~\cite{mvitv2} & Sup., IN-21K & \color{demphcolor}{87.5} & \color{demphcolor}{97.8} & 2828\x3\x4 & 218 \\
MViT-L\textuparrow312, 40\x3~\cite{mvitv2} & \textbf{MaskFeat}, K600 & \textbf{88.3} & \textbf{98.0} & 2828\x3\x4 & 218 \\
\end{tabular}
}
}

\resizebox{1.02\linewidth}{!}{
\subfloat[\textbf{Kinetics-700}\label{tab:k700-finetune}]{
\hspace*{-7pt}
\begin{tabular}{l|l|c|c|r|r}
model & pre-train & top-1 & top-5 & \footnotesize{FLOPs}\x views & \footnotesize{Param} \\
\shline \hline
{SlowFast} {\scriptsize{16\x 8 +NL}}~\cite{slowfast} &\multicolumn{1}{c|}{-} & 71.0 & 89.6 & 234\x3\x10 & 60 \\
MoViNet-A6~\cite{movinet} & \multicolumn{1}{c|}{-} & 72.3 & N/A & 386\x1\x1 & 31 \\
\shline\hline
MViT-L, 16\x4~\cite{mvitv2} & Sup., IN-21K & \color{demphcolor}{76.7} & \color{demphcolor}{93.4} & 377\x1\x10 & 218 \\
MViT-L, 16\x4~\cite{mvitv2} & \textbf{MaskFeat}, K700 & \textbf{77.5} & \textbf{93.8} & 377\x1\x10 & 218 \\
\hline
MViT-L\textuparrow312, 40\x3~\cite{mvitv2} & Sup., IN-21K & \color{demphcolor}{79.4} & \color{demphcolor}{94.9} & 2828\x3\x4 & 218 \\
MViT-L\textuparrow312, 40\x3~\cite{mvitv2} & \textbf{MaskFeat}, K700 & \textbf{80.4} & \textbf{95.7} & 2828\x3\x4 & 218 \\
\end{tabular}
}
}
\vspace{-10pt}
\caption{\textbf{Comparison with previous work on K600 \& K700}.}
\label{tab:kinetics-finetune}
\vspace{-5pt}
\end{table}

\paragraph{Kinetics-600 and Kinetics-700.}
\cref{tab:kinetics-finetune} compares with prior work on K600~\cite{k600} and K700~\cite{k700}. Both are larger versions of Kinetics. An \mbox{MViT-L, 16\x4} is pre-trained with MaskFeat for 300 epochs and fine-tune for 75 epochs on both datasets. The models achieve the top accuracy of {86.4}\% on K600 and {77.5}\% on K700, using no external image data amd over 10\x fewer FLOPs compared to previous Transformer-based methods. 

Finally, we fine-tune these \mbox{MViT-L, 16\x4} models at a larger input resolution of 312 and a longer duration of 40\x3 to achieve \textbf{88.3}\% top-1 on K600 and \textbf{80.4}\% top-1 on K700, setting a new state-of-the-art with a large margin over the previous best approaches on each dataset, \textit{without} any external supervised pre-training (\eg on IN-21K or JFT-300M).

\subsection{Transfer Learning}
We evaluate transfer learning in downstream tasks using the \mbox{MViT-L\textuparrow312, 40\x3} Kinetics models in \cref{tab:k400-finetune,tab:k600-finetune}.

\begin{table}[t!]
\centering
\vspace{-1pt}
\hspace*{-7pt}
\tablestyle{2.0pt}{1.04}
\resizebox{1.02\linewidth}{!}{
\begin{tabular}{l|c|cc|r|r}
{model} & pre-train & {\footnotesize center} & {\footnotesize full} & {\scriptsize FLOPs} & {\scriptsize Param} \\ 
\shline\hline
SlowFast R101, 8\x8~\cite{slowfast} & \multirow{2}{*}{K400} & 23.8 & - & 138 & 53 \\
MViTv1-B, 64\x3~\cite{mvit} & & 27.3 & - & 455 & 36 \\
\hline
{SlowFast} 16\x8 +NL~\cite{slowfast} & \multirow{6}{*}{K600} & 27.5 & - & 296 & 59 \\
X3D-XL~\cite{x3d}  & & 27.4 & - & 48 & 11 \\
MViTv1-B-24, 32\x3~\cite{mvit}  & & 28.7 & - & 236 & 53 \\
Object Transformer~\cite{object-transformer} & & 31.0 & - & 244 & 86\\
ACAR R101, 8\x8 +NL~\cite{acar} & & - & 31.4 & N/A & N/A \\ \hline
ACAR R101, 8\x8 +NL~\cite{acar} & K700 & - & 33.3 & N/A & N/A \\
\shline\hline
MViT-L{\scriptsize \textuparrow312, 40\x3}~\cite{mvitv2}, Sup. & {\footnotesize IN-21K+K400}  & 31.6 & - & 2828 & 218 \\ 
MViT-L{\scriptsize\textuparrow312, 40\x3}~\cite{mvitv2}, \textbf{MaskFeat} & K400 & 37.3 & 38.5 & 2828 & 218 \\ 
\hline
MViT-L{\scriptsize\textuparrow312, 40\x3}~\cite{mvitv2}, \textbf{MaskFeat} & K600 & \textbf{38.8} & \textbf{39.8} & 2828 & 218 \\ 
\end{tabular}
}
\vspace{-9pt}
\caption{\textbf{Transferring to AVA v2.2}~\cite{ava}. We use single center crop inference (\textit{center}) following MViTv1~\cite{mvit} and full resolution inference (\textit{full}) to compare to the 2020 AVA Challenge winner ACAR~\cite{acar}. Inference cost is with the \textit{center} strategy.}
\label{tab:ava}
\vspace{-10pt}
\end{table}

\paragraph{Action detection.}
AVA v2.2~\cite{ava} is a benchmark for spatiotemporal localization of human actions. We fine-tune the \mbox{MViT-L\textuparrow312, 40\x3} Kinetics models on AVA v2.2. Details are in \S\ref{app:detail-ava}. \cref{tab:ava} reports mean Average Precision (mAP) of our MaskFeat models compared with prior state-of-the-art.
MaskFeat only using K400 contributes to a significant gain of +\textbf{5.7} mAP over its IN-21K pre-trained counterpart using \textit{identical} architectures. By utilizing a larger video dataset, K600, the model reaches an unprecedented accuracy of \textbf{39.8} mAP with full resolution testing, \textit{greatly surpassing all previous methods}, including ActivityNet challenge winners. The strong performance of MaskFeat on AVA suggests a clear advantage of \textit{masked modeling on video} over \textit{supervised classification on image} pre-training for this localization-sensitive recognition task.

\begin{table}[t!]
\centering
\hspace*{-7pt}
\tablestyle{2.0pt}{1.04}
\resizebox{1.02\linewidth}{!}{
\begin{tabular}{l|c|c|c|r|r}
{model} & pre-train & top-1  & top-5 & {\scriptsize FLOPs} & {\scriptsize Param} \\ 
\shline\hline
SlowFast, R101, 8\x8~\cite{slowfast} & \multirow{2}{*}{K400} & 63.1 & 87.6 & 106 & 53 \\
MViTv1-B, 64\x3~\cite{mvit} & & 67.7 & 90.9 & 455 & 37 \\
\hline
MViTv1-B-24, 32\x3~\cite{mvit} & K600 & 68.7 & 91.5 & 236 & 53.2 \\
\hline
Mformer-L~\cite{mformer} & \multirow{3}{*}{\footnotesize IN-21K+K400} & 68.1 & 91.2 & 1185 & 109 \\
ORViT Mformer-L~\cite{orvit} & & 69.5 & 91.5 & 1259 & 148 \\
Swin-B, 32\x3~\cite{video-swin} & & 69.6 & 92.7 & 321 & 89 \\
\shline\hline
MViT-L{\scriptsize \textuparrow312, 40\x3}~\cite{mvitv2}, Sup. & {\footnotesize IN-21K+K400} & 73.3 & 94.1 & 2828 & 218 \\ 
MViT-L{\scriptsize\textuparrow312, 40\x3}~\cite{mvitv2}, \textbf{MaskFeat} & K400 & 74.4 & 94.6 & 2828 & 218 \\ 
\hline
MViT-L{\scriptsize\textuparrow312, 40\x3}~\cite{mvitv2}, \textbf{MaskFeat} & K600 & \textbf{75.0} & \textbf{95.0} & 2828 & 218 \\ 
\end{tabular}
}
\vspace{-10pt}
\caption{\textbf{Transferring to Something-Something v2}~\cite{ssv2}. We report FLOPs with a single ``view''. All entries use one temporal clip and three spatial crops (inference cost is FLOPs\x3\x1).}
\vspace{-10pt}
\label{tab:ssv2}
\end{table}

\paragraph{Human-object interaction classification.}
We fine-tune the \mbox{MViT-L\textuparrow312, 40\x3} Kinetics models in \cref{tab:k400-finetune,tab:k600-finetune} to Something-Something v2 (SSv2)~\cite{ssv2} which focuses on human-object interaction classification. \cref{tab:ssv2} presents the results and details are in \cref{app:detail-ssv2}. In contrast to Kinetics, SSv2 requires fine-grained motion distinctions and temporal modeling to distinguish interactions like \textit{picking something up} and \textit{putting something down}.

Despite the differences between the \textit{supervised tasks} of Kinetics and SSv2, pre-training on Kinetics \textit{without supervised labels} using MaskFeat still contributes to a large gain on fine-tuning accuracy of SSv2. Specifically, MaskFeat with only K400 data contributes to +1.1\% top-1 over its IN-21K+K400 pre-trained counterpart. By utilizing the larger K600, the model reaches an unprecedented \textbf{75.0}\% top-1 accuracy, surpassing all previous methods. This suggests that MaskFeat can learn \textit{spatiotemporal representations} from unlabeled Kinetics data which is known as \textit{appearance-biased}, through self-supervised masked feature prediction.

\subsection{Ablations for Video Recognition}

The ablations are with \mbox{MViT-S, 16\x4} pre-trained for 300 epochs on K400 if not specified. We report 200-epoch fine-tuning accuracy (\%) on K400.

\paragraph{Masking strategy.}
\label{sec:masking-strategy}
We study the masking strategy for spatiotemporal video data. In video, tokens sharing the same spatial position usually also share visual patterns. Therefore, we explore how to handle this redundancy brought by the addition of the temporal dimension.  We consider three different ways of masking and present the results in \cref{tab:vid-mask-strategy}. All entries share the same 40\% masking ratio.

\begin{table}[!ht]
\vspace{-5pt}
\centering
\tablestyle{5.0pt}{1.05}
\begin{tabular}{r|x{40}x{40}>{\columncolor{defaultcolor}}x{40}}
masking  &  frame &tube  & cube \\
\shline
top-1  &  81.0\macc{-1.2} & 81.9\macc{-0.3}  & \textbf{82.2}  \\
\end{tabular}
\vspace{-10pt}
\caption{\textbf{Masking strategy.} Varying the strategy of masking in spatiotemporal data. The default entry is highlighted in \colorbox{defaultcolor}{gray}.}
\label{tab:vid-mask-strategy}
\vspace{-13pt}
\end{table}

First, we consider ``\textit{frame}'' masking, which \textit{independently} masks out consecutive frames. This strategy mostly masks \textit{different} spatial blocks in consecutive frames, but the model could temporally ``interpolate'' between frames to solve the task. This strategy only obtains 81.0\% top-1. 

Second, we consider ``\textit{tube}'' masking. Namely, we first sample a 2-D mask map by block-wise masking as for images, and then extend the 2-D map by repeating it in the temporal dimension. Thus, the masked area is a \textit{straight tube} in a video clip, in which the spatially masked area is the same for every frame. \textit{Tube} masking refrains from relying on the temporal repetition to predict the masked content in static video. It leads to 81.9\% accuracy. 

Third, we consider ``\textit{cube}'' masking, which includes both spatial and temporal blocks that are masked out. This is achieved by sampling random  ``\textit{cubes}'' of tokens until a certain masking ratio is reached. Cubes are sampled by first creating a 2-D block at a random time step, then extending in the temporal dimension with a \textit{random} number of consecutive frames. Therefore, \textit{cube} masking can be considered as an generalization of \textit{tube} and \textit{frame} masking.  It produces 82.2\% accuracy when used for pre-training. 

Overall, the results in \cref{tab:vid-mask-strategy} show that 
\textit{cube} masking performs best, suggesting both spatial and temporal cues are helpful in masked spatiotemporal prediction.

\begin{table}[!ht]
\centering
\tablestyle{5pt}{1.05}
\begin{tabular}{z{30}|>{\columncolor{defaultcolor}}x{70}x{70}}
type  & center patch  & cube \\
\shline
top-1 & \textbf{82.2} & 82.0\macc{-0.2} \\
\end{tabular}
\vspace{-10pt}
\caption{\textbf{Target design.} Predicting \textit{center patch} HOG or all HOG in a \textit{cube} gives similar results. Default in \colorbox{defaultcolor}{gray}.}
\label{tab:vid-target}
\vspace{-13pt}
\end{table}

\paragraph{Target design.}
\label{app:video-target}
On video, each output token corresponds to a space-time cube. Our default setting is to simply predict the feature of the 2-D spatial patch temporally centered in each masked space-time cube. In \cref{tab:vid-target} we consider another straightforward way of predicting the entire cube, \ie, HOG features of each 2-D patch in the 3-D cube. Results are similar and we use center patch prediction for simplicity.

\begin{table}[!ht]
\centering
\vspace{-5pt}
\tablestyle{5pt}{1.05}
\begin{tabular}{r|x{40}>{\columncolor{defaultcolor}}x{40}x{40}x{40}}
ratio    & 20\%            & 40\%          & 60\%            & 80\% \\
\shline
top-1    & 81.9\macc{-0.3} & \textbf{82.2} & \textbf{82.2} & 82.0\macc{-0.2} \\
\end{tabular}
\vspace{-10pt}
\caption{\textbf{Masking ratio.} Varying  the  percentage  of masked  patches. MaskFeat is robust to masking ratio in video domain.}
\label{tab:vid-mask-ratio}
\vspace{-13pt}
\end{table}

\paragraph{Masking ratio.}
We study the effect of the masking ratio in \cref{tab:vid-mask-ratio}. Interestingly, a \textit{wide} range of masking ratios from 40\% to the extreme 80\% can produce similar fine-tuning accuracy, and only a small ratio of 20\% leads to a slight drop of -0.3\%. This is different from the observation on images, where ratios larger than 40\% lead to degraded accuracy (see discussions in \cref{app:mask-ratio-img}). This indicates that in the video domain visual patterns are indeed more \textit{redundant} than in images, and thus MaskFeat enjoys a larger masking ratio to create a properly difficult task.

\begin{table}[ht!]
\vspace{-5pt}
\centering
\tablestyle{7pt}{1.02}
\begin{tabular}{cc|x{38}x{38}}
epoch           & param. (M) & 300  & 800 \\
\shline
MViT-S, 16\x4   & 36         & 82.2 & 82.0\macc{-0.2} \\
MViT-L, 16\x4   & 218        & 83.1 & 84.3\pacc{+1.2} \\
\end{tabular}
\vspace{-10pt}
\caption{\textbf{Pre-training schedule.} Large model benefits more from longer pre-training schedule.}
\label{tab:video-schedule}
\vspace{-13pt}
\end{table}

\paragraph{Pre-training schedule.}
\label{sec:video-schedule}
We show different pre-training schedule lengths on K400 in \cref{tab:video-schedule}. Each result is fine-tuned from a fully trained model instead of an intermediate checkpoint. For MViT-S with 36M parameters, extending pre-training from 300 epochs to 800 epochs results in a small performance degradation of 0.2\% accuracy. In contrast, for MViT-L longer pre-training provides a significant gain of +1.2\% accuracy. This suggests that MaskFeat is a scalable pre-training task that can be better utilized by models with larger capacity and longer schedule.

\section{Experiments: Image Recognition}

\paragraph{Settings.} 
The evaluation protocol is pre-training followed by end-to-end fine-tuning. We use vanilla base and large models in ViT~\cite{vit} without modification. Our models are pre-trained at 224$^2$ resolution on IN-1K~\cite{imagenet} training set without labels. We use minimal data augmentation: random resized cropping and horizontal flipping. We randomly mask out 40\% of total image patches with block-wise masking following BEiT~\cite{beit}.
More details are in \cref{app:detail-cls}.

\begin{table}[t!]
\centering
\hspace*{-7pt}
\resizebox{1.02\linewidth}{!}{
\tablestyle{5pt}{1.02}
\begin{tabular}{lcccc}
pre-train                                & extra data  & extra model    & ViT-B          & ViT-L \\
\shline
{scratch~\cite{deit}}                & {-}         & -              & {81.8}         & 81.5 \\
supervised\textsubscript{384}~\cite{vit} & IN-21K      & -              & 84.0           & 85.2 \\
MoCo v3~\cite{mocov3}                    & -           & momentum ViT   & 83.2           & 84.1 \\
DINO~\cite{dino}                         & -           & momentum ViT   & 82.8           & -   \\
BEiT~\cite{beit}                         & DALL-E      & dVAE   & 83.2           & 85.2 \\
\shline
\textbf{MaskFeat} (w/ HOG)                    & -           & -             & \textbf{84.0}  & \textbf{85.7} \\ 
\end{tabular}
}
\vspace{-10pt}
\caption{\textbf{Comparison with previous work on IN-1K.} All entries are pre-trained on IN-1K train split, except supervised\textsubscript{384} using IN-21K. MoCo v3 and DINO use momentum encoder. BEiT uses 250M DALL-E data to pre-train dVAE. All entries are trained and evaluated at image size 224$^2$ except supervised\textsubscript{384} at 384$^2$.}
\label{tab:IN-1K}
\vspace{-10pt}
\end{table}

\subsection{Main Results on ImageNet-1K}
\label{sec:exp-img}

In \cref{tab:IN-1K} we compare MaskFeat to previous work including from-scratch, IN-21K supervised pre-training, and previous self-supervised methods. We pre-train MaskFeat for 1600 epochs here while for 300 epochs in \cref{tab:feat-img}. The fine-tuning schedule is the same everywhere and rather short, 100 epochs for ViT-B and 50 epochs for ViT-L.

We observe that MaskFeat pre-training significantly boosts the scratch baselines for both ViT-B and ViT-L. Our approach at image size 224$^2$ is on par with (ViT-B), or even outperforms (ViT-L) supervised pre-training on IN-21K that has 10\x more images and labels at image size 384$^2$. It has been shown~\cite{vit} that ViT models are data-hungry and require large-scale supervised pre-training, possibly due to the lack of typical CNN inductive biases. Our results suggest that MaskFeat pre-training can overcome this without external labeled data by solving our feature inpainting task. Interestingly, more gains are observed on ViT-L compared with ViT-B, suggesting that it is scalable to larger models.

Compared to self-supervised pre-training approaches, MaskFeat is more accurate and simpler. MoCo v3~\cite{mocov3} and DINO~\cite{dino} are contrastive methods that require multi-view training and carefully designed augmentation, while MaskFeat only uses single-views and minimal augmentation. See \cref{app:aug} for ablation on data augmentation of MaskFeat. Compared with BEiT~\cite{beit}, MaskFeat gets rid of the dVAE tokenizer, which introduces both an extra pre-training stage on the 250M DALL-E dataset, and a non-negligible inference overhead during masked prediction. While MaskFeat simply calculates HOG features.

MaskFeat in \cref{tab:IN-1K} is pre-trained for 1600 epochs with a single 224$^2$ view. DINO uses multiple global-local views and an extra momentum encoder, leading to 1535 effective epochs$^\dagger$ (\cref{tab:feat-img}). MoCo v3 saturates after 600 effective epochs~\cite{mocov3}. BEiT is pre-trained for 800 epochs on IN-1K but requires another 1199 effective epochs for dVAE.

We also train the best model in \cref{tab:feat-img}, MaskFeat w/ DINO, for 1600 epochs and it reaches 84.2\%; however, this uses a separate ViT-B model that is trained with another \raisebox{0.5ex}{\texttildelow}{1535} effective epochs using DINO. MaskFeat w/ HOG can reach 84.0\% without extra model.

\subsection{Ablations for Image Recognition}
\label{sec:ablations}

We ablate the design choices of MaskFeat in the image domain first. We use ViT-B pre-trained for 300 epochs by default and report fine-tuning top-1 accuracy (\%) on IN-1K. More ablations (\eg on training epochs) are  in \cref{app:ablation-image}.

\begin{table}[t!]
\centering
\tablestyle{4.5pt}{1.04}
\subfloat[\textbf{Contrast normalization.}\label{subtab:norm}]{
  \begin{tabular}{r|cc>{\columncolor{defaultcolor}}c}
  norm.  & none & $\ell_1$ & $\ell_2$ \\
  \shline
  top-1  & 82.2 & 82.8     & \textbf{83.6} \\
  \end{tabular}
}
\subfloat[\textbf{Color channel.}\label{subtab:channel}]{
  \begin{tabular}{r|c>{\columncolor{defaultcolor}}cc}
  channel & gray & rgb           & opp. \\
  \shline
  top-1   & 83.2 & \textbf{83.6} & 83.5 \\
  \end{tabular}
}
\vspace{5pt}
\subfloat[\textbf{Orientation bins.}\label{subtab:nbins}]{
  \begin{tabular}{r|c>{\columncolor{defaultcolor}}cc}
  \#bins & 6    & 9             & 12 \\
  \shline
  top-1  & 83.4 & \textbf{83.6} & 83.5 \\
  \end{tabular}
}
\subfloat[\textbf{Spatial cell size.}\label{subtab:cellsize}]{
  \begin{tabular}{r|c>{\columncolor{defaultcolor}}cc}
  cell size & 4\x4 & 8\x8          & 16\x16 \\
  \shline
  top-1     & 83.2 & \textbf{83.6} & 83.2 \\
  \end{tabular}
}
\vspace{-5pt}
\caption{\textbf{HOG implementation.} \subref{subtab:norm} Local contrast normalization plays a key role, and \subref{subtab:channel} MaskFeat benefits from color information; this is in line with HOG/SIFT studies on image recognition~\cite{hog,chatfield2014return}. HOG as target is \subref{subtab:nbins} robust to the number of orientation bins, and \subref{subtab:cellsize} benefits from $8\times8$ spatial cell. Opp. represents opponent color space~\cite{colorhog}. Default entries are marked as \colorbox{defaultcolor}{gray}.}
\label{tab:img-hog}
\vspace{-15pt}
\end{table}

\paragraph{HOG implementation.}
We ablate HOG implementation details in \cref{tab:img-hog}. We first investigate the local
contrast normalization in HOG, which is key to its performance in image recognition~\cite{hog}. It is applied by normalizing each histogrammed vector of local 8\x8 pixel cells, which leads \eg to local invariance in illumination change. We show in \cref{subtab:norm} that normalization is essential for MaskFeat. Compared with default $\ell_2$ normalization, using $\ell_1$ normalization results in a 0.8\% drop and \textit{not using any normalization} causes a large -\textbf{1.4}\% drop. Similar results are reported in \cite{hog} for directly using HOG for image recognition.

We next investigate the effectiveness of color information in \cref{subtab:channel}. \textit{Gray} corresponds to HOG on gray-scale images, which only contains intensity information. To include color information in HOG, \textit{rbg} calculates separate gradients for each color channel and concatenates the three histograms. And \textit{opp.} is an affine transformation of RGB to an opponent color space~\cite{colorhog}.  Results show that using color information provides a small gain of around +0.4\% compared with only using gray-scaled intensity information.

We vary the number of orientation and spatial bins in \cref{subtab:nbins} and \cref{subtab:cellsize}, which provides geometric invariance in HOG descriptors. Following HOG\cite{hog}, we use 9 orientation bins that are evenly spaced from 0\textdegree{ }to 180\textdegree{ }(\textit{unsigned}), and use 8$\times$8 pixel cells (SIFT~\cite{sift} uses 8 bins in 8$\times$8 cells). We observe that these default settings in \cite{hog} are good for MaskFeat and that it is robust to different numbers of orientation bins, but a specific size of  8$\times$8 pixels in a cell is the best.

\begin{figure}[t!]
\centering
\captionsetup[subfigure]{labelformat=empty, justification=raggedright}
\subfloat{\begin{adjustbox}{max width=0.98\linewidth}\begin{tabular}{cccc}
\enspace\footnotesize{masked input} & \enspace\footnotesize{pixel prediction} & \enspace\footnotesize{HOG prediction} & \enspace\footnotesize{original image}
\end{tabular}\end{adjustbox}}

\subfloat[\scriptsize{Both two predictions make good sense given a small visible region at the bird's head.}\label{fig:pix-hog-a}]{
\includegraphics[width=0.98\linewidth]{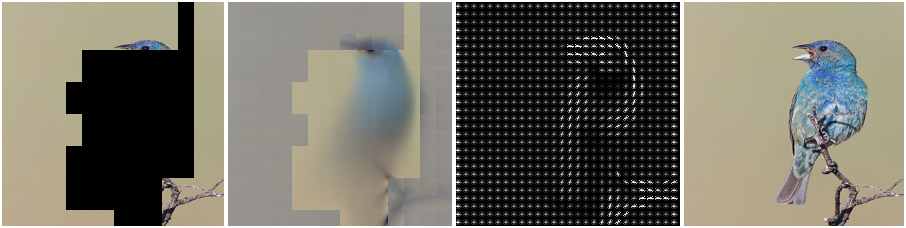}
}

\subfloat[\scriptsize{Pixel with \textbf{color ambiguity}: Though pixel prediction makes a sensible guess on the balloon, the loss penalty is large because of unmatched color (red \vs black).}\label{fig:pix-hog-b}]{
\includegraphics[width=0.98\linewidth]{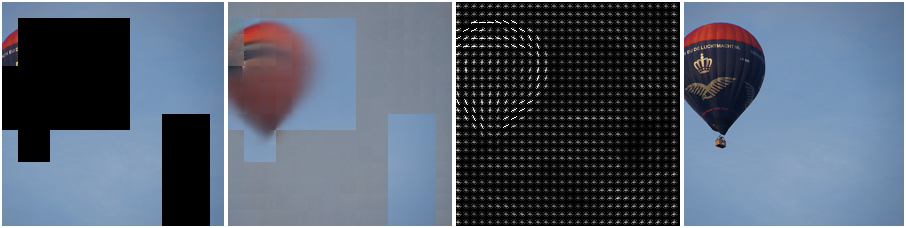}
}

\subfloat[\scriptsize{Pixel with \textbf{texture ambiguity}: Pixel prediction is blurry in texture-rich area because of ambiguity, while HOG successfully characterizes major edge directions.}\label{fig:pix-hog-c}]{
\includegraphics[width=0.98\linewidth]{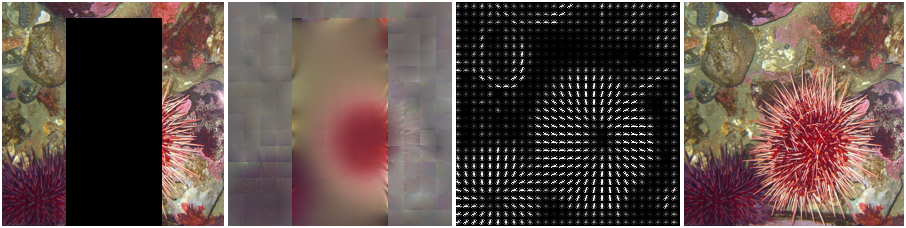}
}
\vspace{-7pt}
\caption{\textbf{Pixel \vs HOG predictions} on IN-1K \textit{validation} images$^\dagger$. Pixel targets can have large errors for \textit{ambiguous} problems, and HOG is more robust to ambiguity by \textit{histogramming} and \textit{normalizing} local gradients. Best viewed in color and zoomed in.}
\footnotesize{$^\dagger$ The unmasked regions are not used for loss and thus qualitatively poor.}
\label{fig:pix-hog}
\vspace{-10pt}
\end{figure}

\paragraph{Pixel \vs HOG.}
We qualitatively compare HOG to pixel colors as the feature target of MaskFeat in \cref{fig:pix-hog}. Both pixel and HOG predictions look reasonable in close proximity to the unmasked input.
However, compared to HOG, pixel color targets come with more \textit{ambiguity}. In the balloon (second) example, the model makes a sensible guess predicting a red balloon, which is black in the original image, resulting in a high loss penalty. In the sea urchin (third) example, the model is just able to make a blurry color-wise guess on the object, which is a natural consequence of minimizing a pixel-wise MSE loss in texture-rich, high-frequency regions~\cite{srgan}. In both cases, HOG reduces the risk of ambiguity: normalizing gradients handles the color ambiguity and spatial binning of gradients texture ambiguity.

\begin{table}[!ht]
\vspace{-5pt}
\centering
\tablestyle{8pt}{1.02}
\begin{tabular}{r|x{40}>{\columncolor{defaultcolor}}x{50}x{50}}
targets & pixel           & HOG           & pixel + HOG \\
\shline
top-1  & 82.5\macc{-1.1} & \textbf{83.6} & 82.3\macc{-1.3} \\ 
\end{tabular}
\vspace{-10pt}
\caption{\textbf{Multi-tasking.} Simply combining two targets with two separate linear prediction heads results in a drop, suggesting conflict in the objectives. The default entry is marked as \colorbox{defaultcolor}{gray}.}
\label{tab:img-multi}
\vspace{-10pt}
\end{table}

\paragraph{Multi-tasking.}
Finally, we investigate if combining different targets in a multi-task loss helps. Specifically, we combine pixel and HOG, two single-stage target features, by predicting each target with a separate linear layer. The two prediction losses are simply averaged with equal weighting. The results are summarized in \cref{tab:img-multi}. We see that multi-tasking of pixel and HOG provides a small gain over the scratch baseline (82.3\% \vs 81.8\%), but the accuracy is lower than pixel or HOG only. Though further tuning the loss weighting might improve this result, it signals that the two objectives can not benefit each other. This is reasonable, as HOG targets are locally normalized while pixel colors are strongly influenced by local brightness changes.

\section{Related Work}
\paragraph{Masked visual prediction} was pioneered with stacked autoencoders~\cite{denoisingae} and inpainting tasks\cite{inpainting} using ConvNets. Since the introduction of ViT~\cite{vit}, masked prediction has re-attracted attention of the vision community, partially inspired by the success of BERT~\cite{bert} in NLP. BERT performs masked language modeling where some input tokens are masked at random and the task it to predict those. BERT pre-trained models scale well and generalize to a wide range of different downstream tasks. 

For vision, different masked prediction objectives have been proposed. iGPT~\cite{igpt} predicts the next pixels of a sequence. ViT~\cite{vit} predicts mean colors of masked patches. BEiT~\cite{beit} and VIMPAC~\cite{vimpac} encode masked patches with discrete variational autoencoder (dVAE)~\cite{vqvae,dall-e}. Masked visual prediction has also been explored in the field of vision-language learning (\eg, \cite{vlbert,vilbert,uniter,vilt}). Compared to BEiT and VIMPAC, our method does not rely on dVAEs but directly regresses specific features of the input.

\paragraph{Self-supervised learning} aims to learn from unlabeled visual data by a pre-text task that is constructed by image/patch operations (\eg, \cite{context-prediction,jigsaw,jigsaw2,colorization,rotation,deepclustering}) and spatiotemporal operations (\eg, \cite{goroshin2015unsupervised,misra2016shuffle,fernando2017self,pathak2017learning,wang2017transitive}). 
Recently, contrastive learning~\cite{dosovitskiy2015discriminative} capitalizes on augmentation invariance. The invariance is achieved by enforcing similarity over distorted views of one image while avoiding model collapse~\cite{instdist,moco,simclr,swav,byol,dino,simsiam,cvrl,videomoco}. The line of contrastive methods learns linearly separable representations, commonly evaluated by linear probing. While in this work we focus on maximizing model performance for the end task with an end-to-end fine-tuning protocol~\cite{beit,vimpac}.

\section{Conclusion}

We present Masked Feature Prediction (MaskFeat), a simple visual pre-training approach that regresses \textit{features} of masked regions. In particular, HOG, a hand-designed feature that was driving visual recognition before the deep learning era, works surprisingly well as the prediction target.
MaskFeat is efficient, generalizes well, and scales to large models for both video and image domains.

Our results are especially groundbreaking for video understanding: There has been a large gap of over 5\% accuracy between supervised pre-training on large-scale image datasets and training-from-scratch methods. MaskFeat has closed this gap by directly pre-training on unlabeled videos. Transfer learning performance is even more impressive where an MaskFeat model surpasses its IN-21K counterpart, which uses 60\x~more labels, by +5.7 mAP on action detection (AVA) and +1.1\% top-1 on human-object interaction recognition (SSv2). These results suggest a clear benefit of masked prediction in the visually richer space-time domain to explore in future work.

\appendix
\section*{Appendix}
In this Appendix, we provide further ablations for image (\S\ref{app:ablation-image}) classification. \S\ref{app:detail} contains the implementation details, and \S\ref{app:ablations-qualitative} provides more qualitative results. 

\section{Ablations on Image Classification}
\label{app:ablation-image}

\begin{table}[!ht]
\vspace{-5pt}
\centering
\tablestyle{8pt}{1.02}
\begin{tabular}{c|x{45}x{45}x{45}}
epoch & 300  & 800             & 1600 \\
\shline
ViT-B & 83.6 & 83.9\pacc{+0.3} & \textbf{84.0}\pacc{+0.4} \\
ViT-L & 84.4 & 85.4\pacc{+1.0} & \textbf{85.7}\pacc{+1.3} \\
\end{tabular}
\vspace{-10pt}
\caption{\textbf{Pre-training schedule.} Gains with longer schedules are observed. The large model benefits more from longer schedules.}
\label{tab:img-schedule}
\vspace{-10pt}
\end{table}

\paragraph{Pre-training schedule.}
We show different lengths of pre-training in \cref{tab:img-schedule}. Each result is fine-tuned from a fully trained model instead of an intermediate checkpoint.

For both base and large size models, improvements are observed with longer pre-training schedules. Interestingly, the large size model benefits more from longer pre-training with +1\% gain from 300 epochs to 800 epochs, while the base-size model is only improved by +0.3\%. This suggests that MaskFeat is a sufficiently difficult task such that (i) excessive long pre-training does not cause over-fitting of large models, and (ii) MaskFeat is sufficiently difficult for high capacity models.
Training for 1600 epochs only gives another +0.1\% improvement for ViT-B.

\begin{table}[!h]
\vspace{-5pt}
\centering
\tablestyle{5pt}{1.02}
\begin{tabular}{r|x{42}>{\columncolor{defaultcolor}}x{42}x{42}x{42}}
ratio & 20\%            & 40\%          & 60\%            & 80\% \\
\shline
top-1 & 83.5\macc{-0.1} & \textbf{83.6} & 83.1\macc{-0.5} & 82.5\macc{-1.1} \\
\end{tabular}
\vspace{-10pt}
\caption{\textbf{Masking ratio (image).} Varying the percentage of masked patches. A smaller percentage of masking is preferred.
}
\vspace{-10pt}
\label{tab:img-mask}
\end{table}

\paragraph{Masking ratio.}
\label{app:mask-ratio-img}
We vary the percentage of masked patches in \cref{tab:img-mask} with block-wise masking following BEiT~\cite{beit}. 
We observe that masking out 20\%\raisebox{0.5ex}{\texttildelow}40\% patches works well and that stronger masking degrades accuracy. MaskFeat requires enough visible patches to set up a meaningful objective. Note that 20\%\raisebox{0.5ex}{\texttildelow}40\% masking is more than 15\% masking used in masked language modeling (BERT~\cite{bert}), reflecting redundancy in raw visual signals.

\begin{table}[!ht]
\centering
\tablestyle{4pt}{1.04}
\subfloat[\textbf{Augmentation.} Our MaskFeat works best with only Random Resized Crop (RRC) as augmentation.\label{subtab:aug}]{
   \begin{tabular}{r|>{\columncolor{defaultcolor}}x{35}x{70}x{70}}
   aug.  & RRC            & RRC + color jit. & RRC + Rand Aug. \\
   \shline
   top-1 & \textbf{83.6}  & \textbf{83.6}    & 83.2\macc{-0.4} \\
   \end{tabular}
}
\vspace{5pt}
\subfloat[\textbf{Random resized crop scale.} A relatively large scale of random crops provides a small gain.\label{subtab:rccscale}]{
  \begin{tabular}{r|x{42}x{42}>{\columncolor{defaultcolor}}x{42}x{42}}
  scale  & $[0.08, 1.0]$   & $[0.2, 1.0]$    & $[0.5, 1.0]$  & $[0.8, 1.0]$ \\
  \shline
  top-1  & 83.4\macc{-0.2} & 83.4\macc{-0.2} & \textbf{83.6} & 83.4\macc{-0.2} \\
  \end{tabular}
}
\vspace{-5pt}
\caption{\textbf{Data augmentation} in MaskFeat. Defaults are  \colorbox{defaultcolor}{gray}.}
\label{tab:img-aug}
\vspace{-10pt}
\end{table}

\paragraph{Data augmentation.}
\label{app:aug}
We study the effect of data augmentation during MaskFeat pre-training in \cref{tab:img-aug}. All three entries in \cref{subtab:aug} use random horizontal flipping. Our approach works best with only random resized crop (RRC), while color jittering has no influence on the result and stronger augmentation (RandAugment~\cite{randaug}) degrades the performance slightly by 0.4\%. This suggests that strong augmentations might lead to artificial patterns that in turn lead to a gap in pre-training and finetuning and MaskFeat works nearly augmentation-free.
Conversely, contrastive-based methods are arguably dependent on ``augmentation engineering'' to provide prior knowledge (\eg, \cite{mocov2,byol}), which could lead to conflicting clues~\cite{purushwalkam2020demystifying} and over-fitting to a specific combination of augmentations~\cite{xiao2020should}.

We further study the effect of the RRC $[\min, \max]$ scales in \cref{subtab:rccscale}. Our approach is robust to this hyper-parameter. MaskFeat works best with low strength of RRC, $[0.5, 1.0]$, which covers a large fraction of each sample.

\begin{table}[!ht]
\vspace{-5pt}
\centering
\tablestyle{8pt}{1.05}
\begin{tabular}{r|x{30}x{30}x{30}}
block              &  \nth{8}       & \nth{16} & \nth{24} \\
\shline
top-1              &  \textbf{67.7} & 66.0     & 55.9 \\
\end{tabular}
\vspace{-10pt}
\caption{\textbf{Linear probing.} We perform linear probing after the \nth{8}, \nth{16}, \nth{24} (last) block of MaskFeat pre-trained ViT-L. Lower layers obtain better linear accuracy.}
\label{tab:img-linear}
\vspace{-10pt}
\end{table}

\paragraph{Linear probing.}
Besides the fine-tuning protocol, we consider linear probing in \cref{tab:img-linear} which is commonly used to evaluate contrastive methods~\cite{moco,simclr}. We train randomly initialized linear classifiers right at transformer block outputs. Specifically, we consider the average pooled outputs of the \nth{8}, \nth{16} and \nth{24} (last) transformer blocks of a ViT-L pre-trained with 1600 epochs of MaskFeat on IN-1K. We observe that lower layers (\eg, the \nth{8}) tend to have higher linear accuracy. This is different from contrastive based methods whose higher layers tend to obtain better linear accuracy~\cite{instdist,cmc}. All layers lag behind contrastive methods by a large margin. For instance, MoCo v3~\cite{mocov3} has 77.6\% at the last block of ViT-L. This suggests that contrastive-based and masked visual prediction methods have very different features. MaskFeat learns good visual knowledge revealed by fine-tuning protocol but not linearly separable features. 

Our hypothesis here is that instance discrimination losses in contrastive learning create different embeddings (classes) for different images which can be largely reduced to class-level information (a subset of classes) with a linear layer. 

\begin{table}[t!]
\subfloat[\textbf{Pre-training setting.}\label{tab:detail_pt}]{
  \tablestyle{3pt}{1.02}
  \begin{tabular}{y{80}|x{65}x{65}}
  config & ImageNet & Kinetics \\
  \shline
  optimizer & \multicolumn{2}{c}{AdamW~\cite{adamw}} \\
  optimizer momentum & \multicolumn{2}{c}{$\beta_1, \beta_2{=}0.9, 0.999$} \\
  weight decay & \multicolumn{2}{c}{0.05} \\
  learning rate schedule & \multicolumn{2}{c}{cosine decay~\cite{sgdr}} \\
  warmup epochs~\cite{1hour} & \multicolumn{2}{c}{30} \\
  augmentation & \multicolumn{2}{c}{hflip, RandomResizedCrop} \\
  gradient clipping & \multicolumn{2}{c}{0.02} \\
  drop path~\cite{droppath} & \multicolumn{2}{c}{\xmark} \\ 
  base learning rate$^\dagger$ & 2e-4 & 8e-4 \\
  batch size & 2048 & 512 \\
  \end{tabular}
}
\vspace{5pt}
\subfloat[\textbf{Fine-tuning setting.}\label{tab:detail_ft}]{
  \tablestyle{3pt}{1.02}
  \begin{tabular}{y{80}|x{29}x{29}x{29}x{29}}
  config & \multicolumn{2}{c}{ImageNet} & \multicolumn{2}{c}{Kinetics} \\
         & \footnotesize{ViT-B} & \footnotesize{ViT-L} & \footnotesize{MViT-S} & \footnotesize{MViT-L} \\
  \shline
  optimizer & \multicolumn{4}{c}{AdamW~\cite{adamw}} \\
  optimizer momentum & \multicolumn{4}{c}{$\beta_1, \beta_2{=}0.9, 0.999$} \\
  weight decay & \multicolumn{4}{c}{0.05} \\
  learning rate schedule & \multicolumn{4}{c}{cosine decay~\cite{sgdr}} \\
  warmup epochs~\cite{1hour} & \multicolumn{4}{c}{5} \\
  augmentation & \multicolumn{4}{c}{RandAug (9, 0.5)~\cite{randaug}} \\
  mixup~\cite{mixup} & \multicolumn{4}{c}{0.8} \\
  cutmix~\cite{cutmix} & \multicolumn{4}{c}{1.0} \\
  label smoothing~\cite{inception} & \multicolumn{4}{c}{0.1} \\
  drop out~\cite{dropout} & \multicolumn{4}{c}{\xmark} \\
  base learning rate$^\dagger$ & 2e-3 & 1e-3 & 4.8e-3 & 9.6e-3 \\
  layer-wise decay~\cite{electra} & 0.65 & 0.75 & 0.75 & 0.875   \\
  batch size & 2048 & 1024 & 512 & 256  \\
  training epochs & 100 & 50 & 200 & 75 \\
  drop path~\cite{droppath} & 0.1 & 0.1 & 0.1 & 0.2 \\
  \end{tabular}
}
\vspace{-5pt}
\caption{\textbf{Configurations for ImageNet and Kinetics.} $^\dagger$We use the linear \textit{lr} scaling rule~\cite{1hour}: \textit{lr} = \textit{base\_lr}\x\textit{batch\_size} / 256.}
\vspace{-10pt}
\end{table}

\section{Implementation Details}
\label{app:detail}

\subsection{ImageNet and Kinetics Experiments}
\label{app:detail-cls}

\paragraph{Architecture.}
For \textit{ImageNet} experiments, we use the standard ViT architecture~\cite{vit} in base and large sizes. We use a single linear layer to transform the output of the last block to form the target predictions. We do not use relative positional bias or layer scaling.

For \textit{Kinetics} experiments, we use the improved MViT architecture~\cite{mvitv2} and we term the original architecture in \cite{mvit} as MViTv1. There are two main modifications. First, instead of using absolute positional embeddings as in MViTv1, relative positional embeddings~\cite{relpos} are incorporated, which are \textit{decomposed} in height, width, and temporal axes. Second, a new residual pooling connection is introduced inside the attention blocks. Specifically, the pooled query tensor is added to the output sequence of self-attention. These two modifications improve the training-from-scratch and supervised-pre-trained baselines. We do not use channel dimension expansion within attention blocks~\cite{mvitv2} but at MLP outputs~\cite{mvit} which has similar accuracy. Our approach which focuses on pre-training techniques is orthogonal to these architectural modifications and provides further gains over the improved baselines.

Unlike ViT models sharing the spatial size of 14$^2$ for all blocks, the MViT architecture is multi-scale and has four scale stages. \textit{Stage 1} output is of spatial size 56$^2$ and \textit{stage 4} output is of spatial size 7$^2$. To share hyper-parameters with ViT models which are of spatial size 14$^2$, we remove MViTs' query pooling before the last MViT stage for MaskFeat pre-training only, resulting in a 14$^2$ final output size, the same as ViT models. This modification introduces little extra computation as \textit{stage 4} is small and has only two Transformer blocks. For the fine-tuning stage, the MViT models are unchanged, with 7$^2$ output to fairly compare with the MViT baselines. Relative positional embeddings are linearly interpolated when the shape is not matched.

When sampling masked tokens for MViT models on the pre-training stage, we first sample a map of the final output size, 14$^2$. This masking map is then nearest-neighbor resized to the \textit{stage 1} size or input size, 56$^2$. In this way the set of input tokens corresponding to the same output token are masked out together, avoiding trivial predictions.  

\paragraph{Pre-training.}
\cref{tab:detail_pt} summarizes the pre-training configurations. Most of the configurations are \textit{shared} by ImageNet and Kinetics, without specific tuning. This shows that MaskFeat is \textit{general} across tasks. The gradient clipping value is set after monitoring training loss over short runs. It is 0.02 for HOG targets and 0.3 for pixel color prediction and deep feature targets.

\paragraph{Fine-tuning.}
\cref{tab:detail_ft} summarizes the fine-tuning configurations. Most of the configurations are \textit{shared} across models, except that deeper models use larger layer-wise learning rate decay and larger drop path rates.

For extra-large, long-term video models with 312 and 352 spatial resolutions as well as 32\x3 and 40\x3 temporal durations, we initialize from their 224 resolution, 16\x4 duration counterparts, disable mixup, and fine-tune for 30 epochs with a learning rate of 1.6e-5 at batch size 128, a weight decay of 1e-8, a drop path~\cite{droppath} rate of 0.75 and a drop out rate of 0.5 for the final linear projection. Other parameters are shared with \cref{tab:detail_ft}.

\subsection{AVA Experiments}
\label{app:detail-ava}

The AVA action detection dataset~\cite{ava} assesses the spatiotemporal localization of human actions in videos. It has 211K training and 57K validation video segments. We evaluate methods on AVA v2.2 and use mean Average Precision (mAP) on 60 classes as is standard in prior work~\cite{slowfast}.

We use \mbox{MViT-L\textuparrow312, 40\x3} as the backbone and follow the same detection architecture in \cite{slowfast,mvit,mvitv2} that adapts Faster R-CNN~\cite{faster-rcnn} for video action detection. Specifically, we extract region-of-interest (RoI) features~\cite{fast-rcnn} by frame-wise RoIAlign~\cite{mask-rcnn} on the spatiotemporal feature maps from the last MViT layer. The RoI features are then max-pooled and fed to a per-class sigmoid classifier for action prediction. The training recipe is identical to \cite{mvit,mvitv2} and summarized next. The region proposals are identical to the ones used in \cite{slowfast,mvit,mvitv2}. We use proposals that have overlaps with ground-truth boxes by IoU $>$ 0.9 for training. The models are trained with synchronized SGD training with a batch size of 64. The base learning rate is 0.6 per 64 batch size with cosine decay~\cite{sgdr}. We train for 30 epochs with linear warm-up~\cite{1hour} for the first five epochs and use a weight decay of 1e-8, a drop path of 0.4 and a head dropout of 0.5.

\subsection{SSv2 Experiments}
\label{app:detail-ssv2}

The SSv2 dataset~\cite{ssv2} contains 169K training, and 25K validation videos with 174 human-object interaction classes. We fine-tune the pre-trained \mbox{MViT-L\textuparrow312, 40\x3} Kinetics models and take the same recipe as in \cite{mvit,mvitv2}. Specifically, we train for 40 epochs with a batch size of 128. The base learning rate is 0.02 per 128 batch size with cosine decay~\cite{sgdr}. We adopt synchronized SGD and use weight decay of 1e-4
and drop path rate of 0.75. The training augmentation is the same as Kinetics in \cref{tab:detail_ft}, except we disable random flipping in training. We use the segment-based input frame sampling~\cite{mvit,tsm} (split each video into segments, and sample one frame from each segment to form a clip). During inference, we take a single temporal clip and three spatial crops over a single video.

\section{Qualitative Experiments}
\label{app:ablations-qualitative}

\begin{figure*}
\centering
\tablestyle{0.2pt}{0.2}
\begin{tabular}{ccc}
  \includegraphics[width=0.33\textwidth]{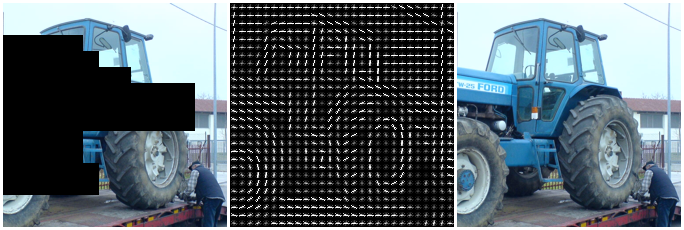} &
  \includegraphics[width=0.33\textwidth]{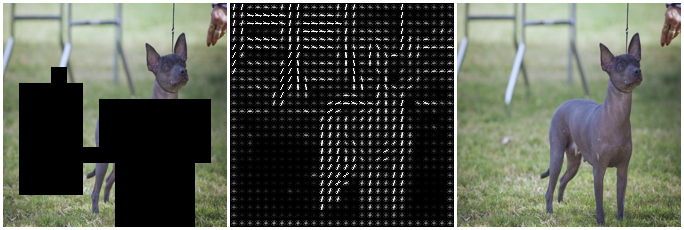} &
  \includegraphics[width=0.33\textwidth]{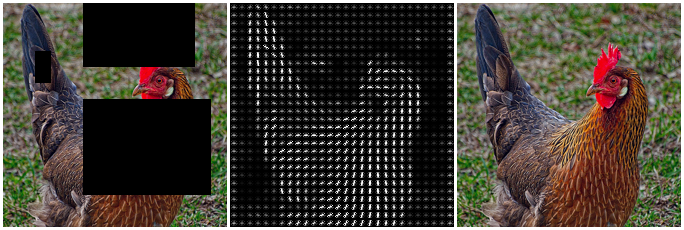} \\
  \includegraphics[width=0.33\textwidth]{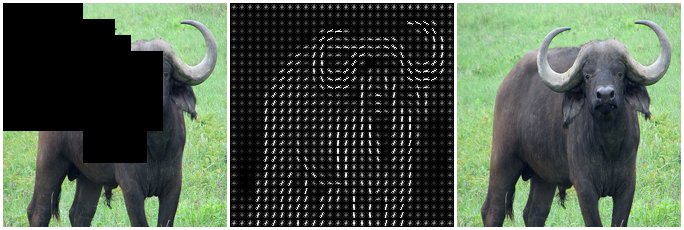} &
  \includegraphics[width=0.33\textwidth]{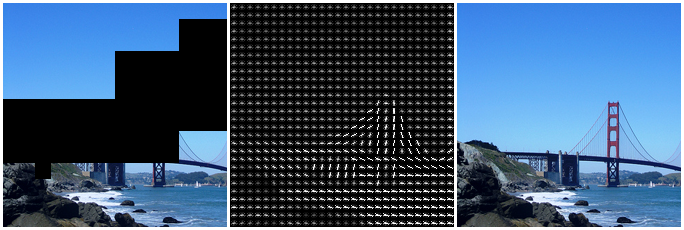} & 
  \includegraphics[width=0.33\textwidth]{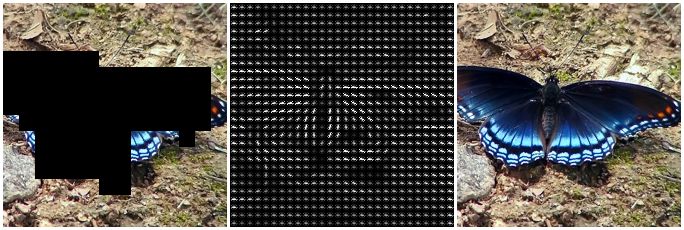} \\
  \includegraphics[width=0.33\textwidth]{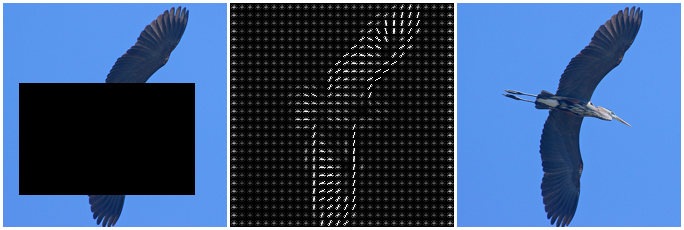} &
  \includegraphics[width=0.33\textwidth]{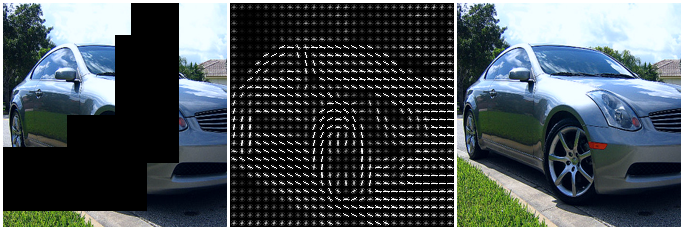} & 
  \includegraphics[width=0.33\textwidth]{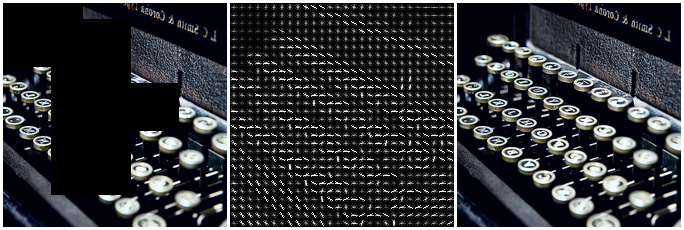} \\
  \includegraphics[width=0.33\textwidth]{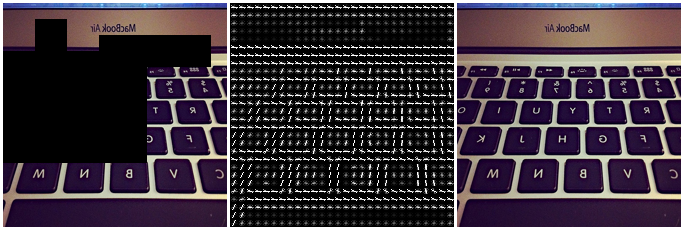} &
  \includegraphics[width=0.33\textwidth]{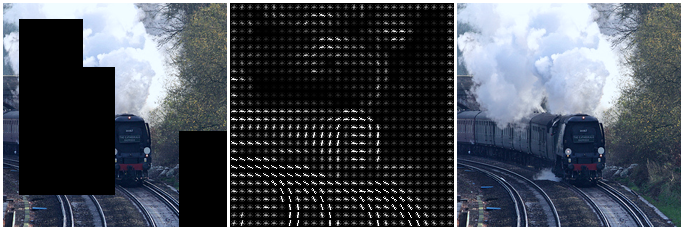} & 
  \includegraphics[width=0.33\textwidth]{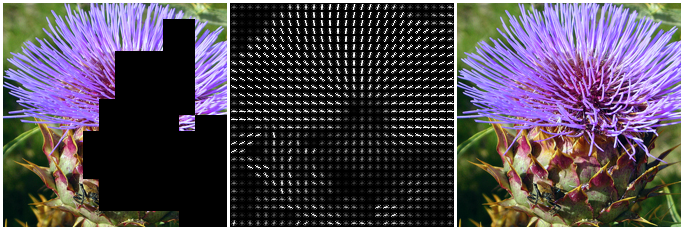} \\
  \includegraphics[width=0.33\textwidth]{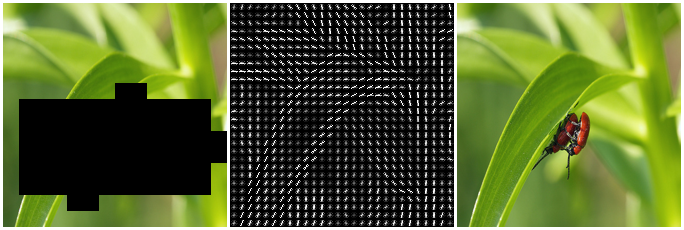} &
  \includegraphics[width=0.33\textwidth]{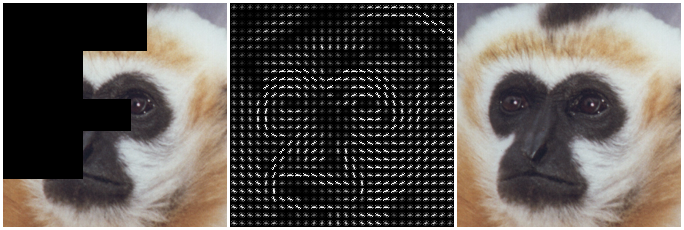} & 
  \includegraphics[width=0.33\textwidth]{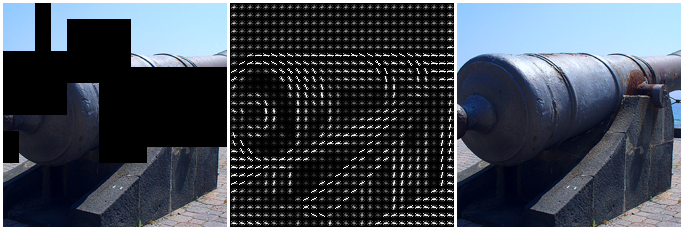} \\
  \includegraphics[width=0.33\textwidth]{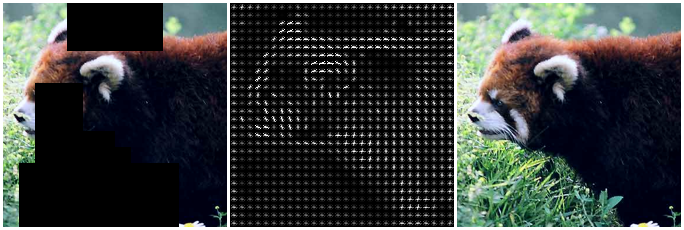} &
  \includegraphics[width=0.33\textwidth]{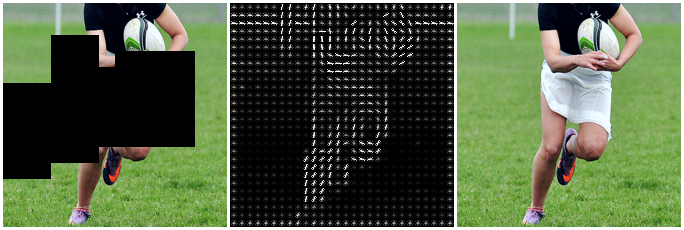} & 
  \includegraphics[width=0.33\textwidth]{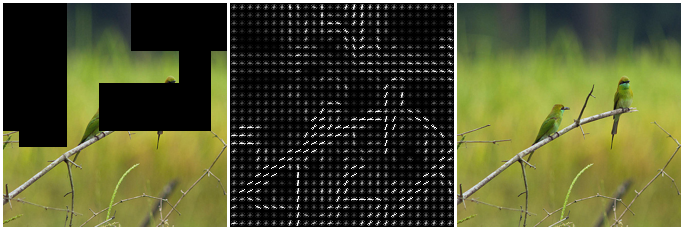} \\
  \includegraphics[width=0.33\textwidth]{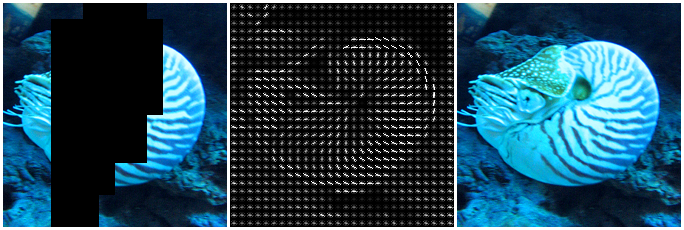} &
  \includegraphics[width=0.33\textwidth]{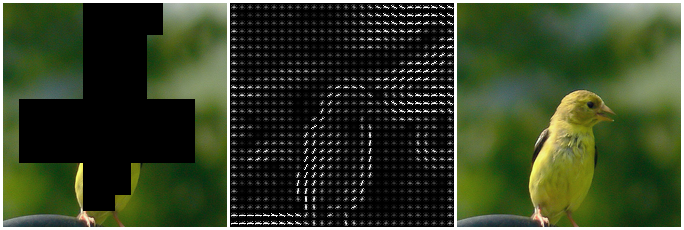} & 
  \includegraphics[width=0.33\textwidth]{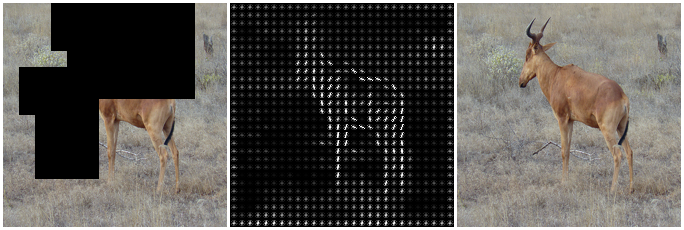} \\
  \includegraphics[width=0.33\textwidth]{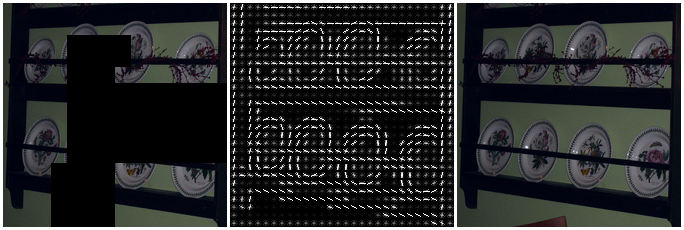} &
  \includegraphics[width=0.33\textwidth]{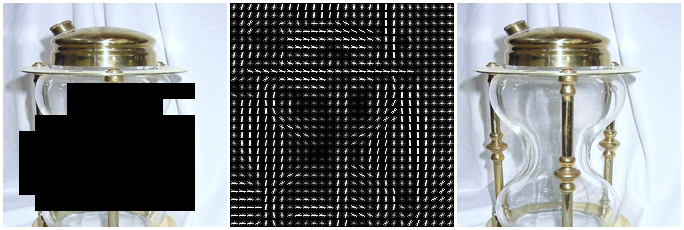} & 
  \includegraphics[width=0.33\textwidth]{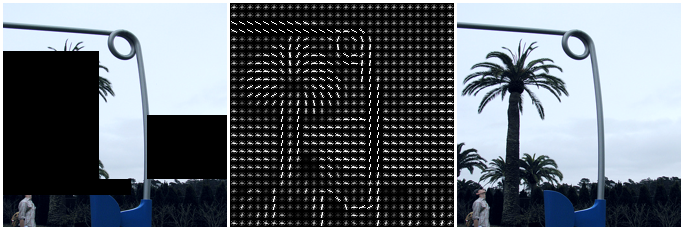} \\
  \includegraphics[width=0.33\textwidth]{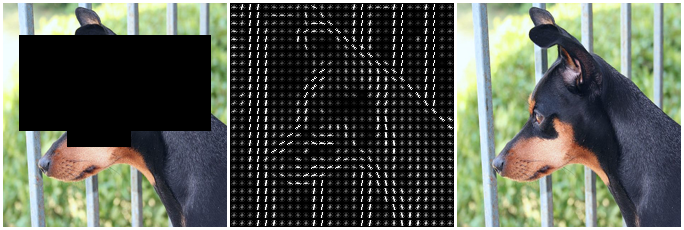} &
  \includegraphics[width=0.33\textwidth]{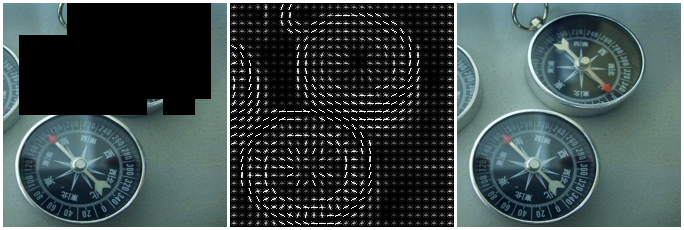} & 
  \includegraphics[width=0.33\textwidth]{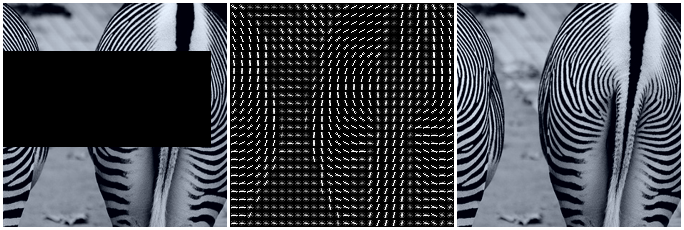} \\
  \includegraphics[width=0.33\textwidth]{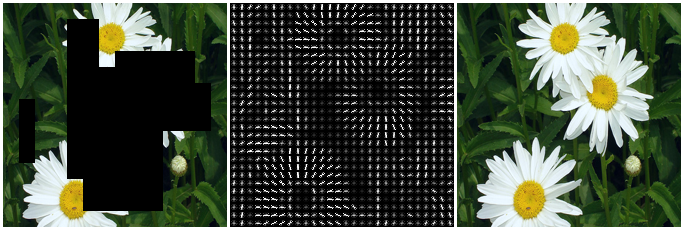} &
  \includegraphics[width=0.33\textwidth]{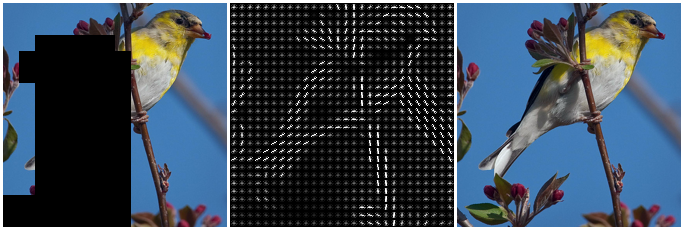} & 
  \includegraphics[width=0.33\textwidth]{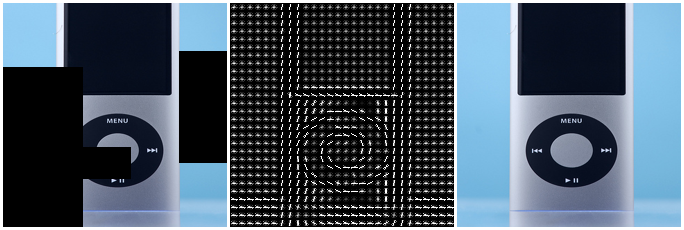} \\
  \includegraphics[width=0.33\textwidth]{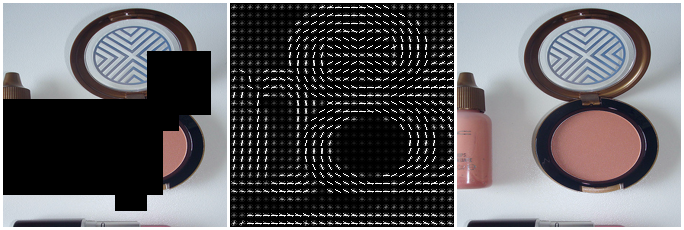} &
  \includegraphics[width=0.33\textwidth]{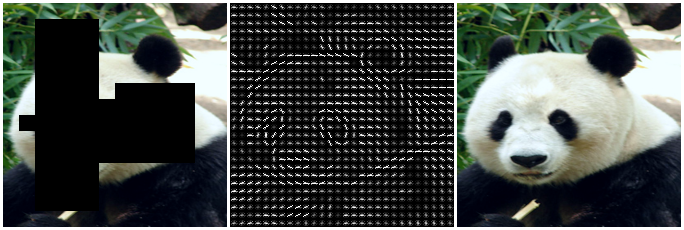} & 
  \includegraphics[width=0.33\textwidth]{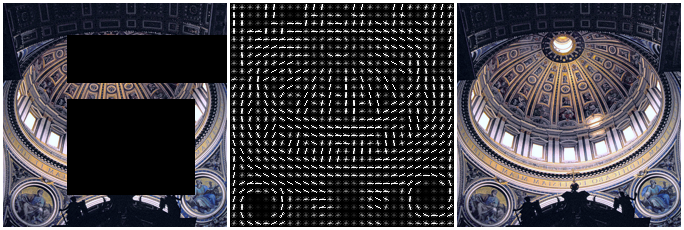} \\
\end{tabular}
\vspace{-5pt}
\caption{More visualizations of HOG predictions. The images are from IN-1K \textit{validation} set. For each column, we show masked input (\textit{left}), HOG predictions (\textit{middle}) and original images (\textit{right}). Original images are not used for prediction. Best viewed in color with zoom. }
\label{fig:more-viz-image}
\end{figure*}

\begin{figure*}
\centering
\tablestyle{0.2pt}{0.2}
\begin{tabular}{ccc}
  \includegraphics[width=0.33\textwidth]{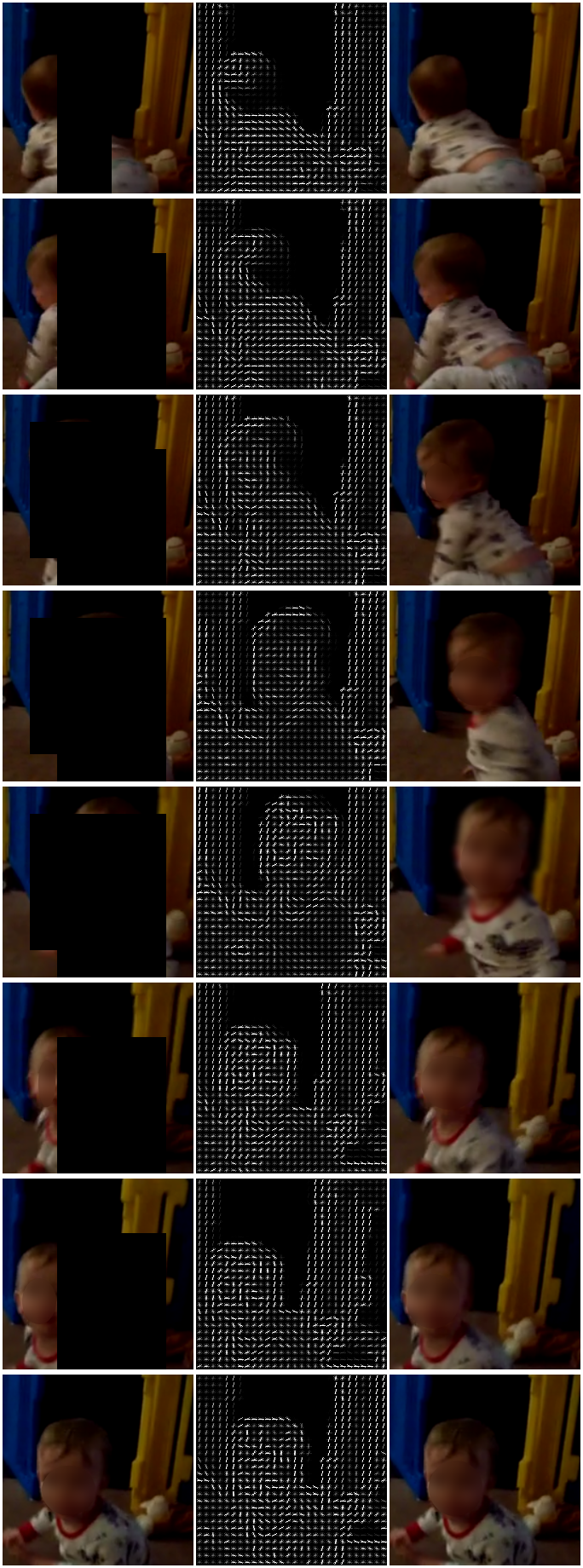} &
  \includegraphics[width=0.33\textwidth]{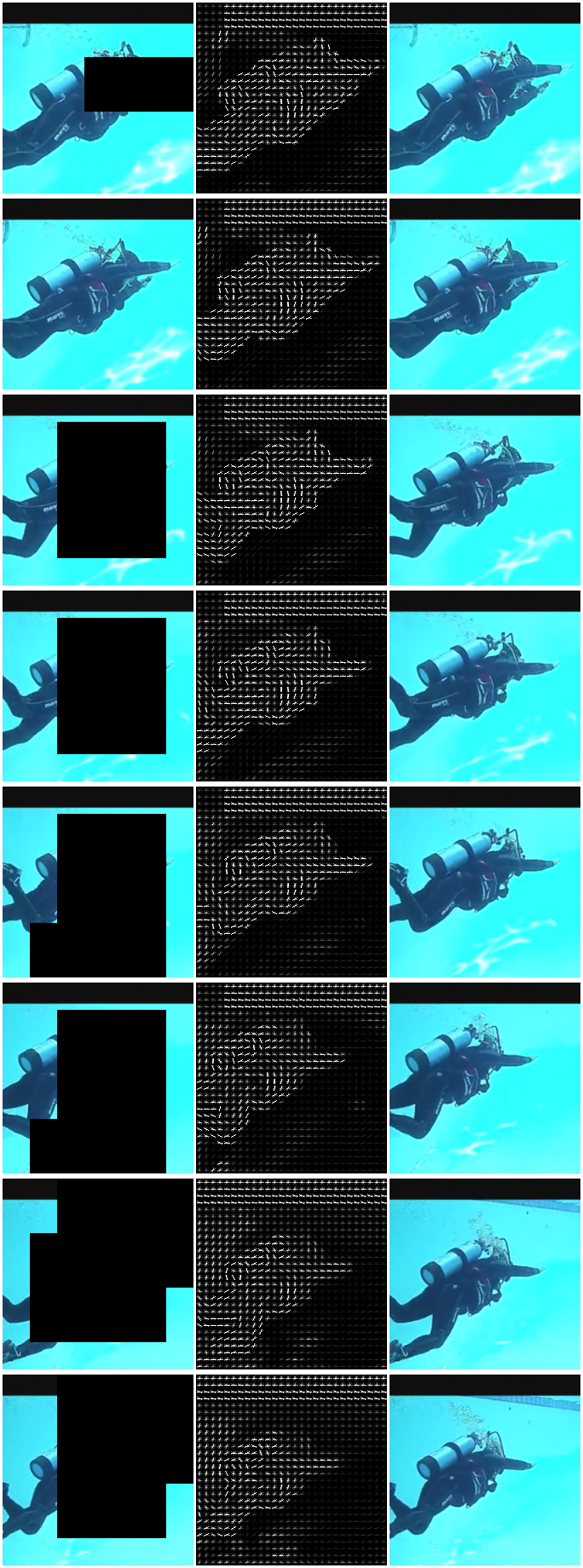} &
  \includegraphics[width=0.33\textwidth]{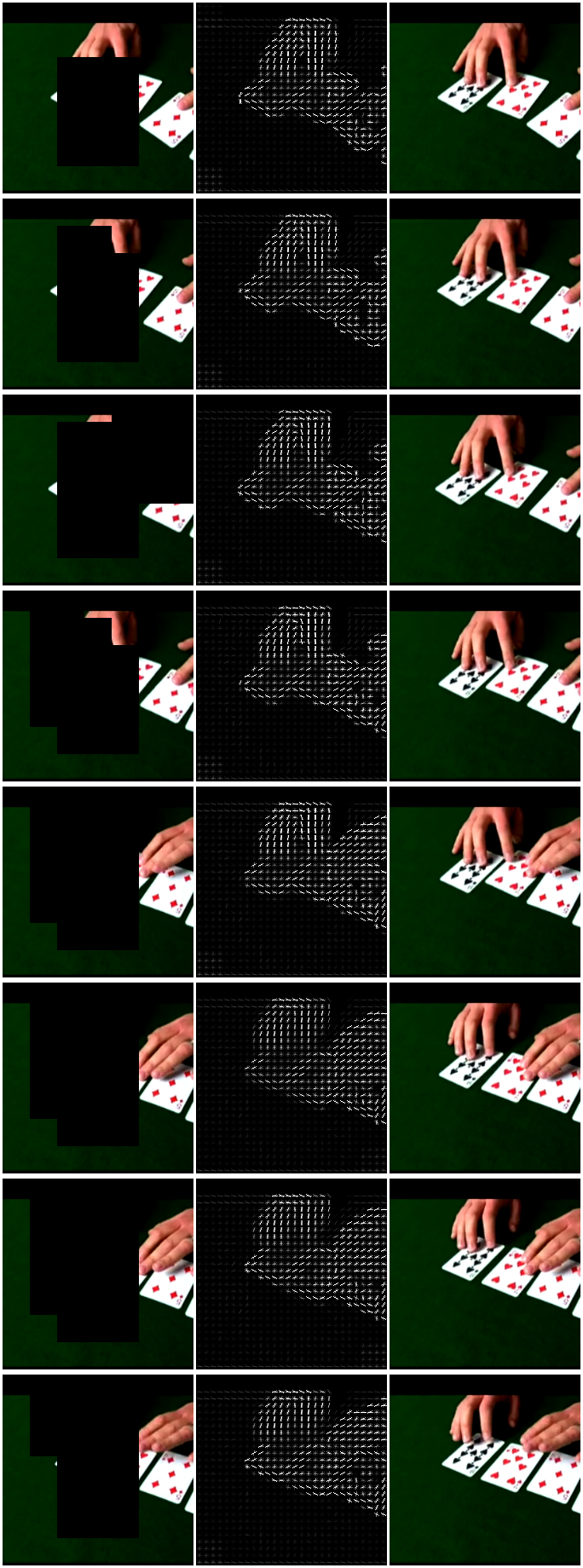} \\
\end{tabular}
\vspace{-5pt}
\caption{More visualizations of HOG predictions (video). The video clips are from K400 \textit{validation} set. For each column, we show masked input (\textit{left}), HOG predictions (\textit{middle}) and original video frames (\textit{right}), and we show eight frames from top to bottom. Original video clips are not used for prediction. Best viewed in color with zoom. }
\label{fig:more-viz-video}
\end{figure*}

We provide more qualitative results of image HOG predictions in \cref{fig:more-viz-image} using ImageNet-1K validation images and for video HOG predictions in \cref{fig:more-viz-video} using Kinetics-400 validation videos.

\section{Acknowledgement}
This project was partially influenced by initial signals of the MAE project~\cite{mae}. We thank Kaiming He and Huiyu Wang for feedback on the manuscript.

\section{Changes in arXiv v2}

We update MaskFeat AVA results in \cref{tab:ava} according to the \href{https://github.com/activitynet/ActivityNet/commit/c070fdb90a70f1260b512ed3f28b23affdeb9641#diff-ee847761e49d609a222e6a7a772f6e02246f18dda3f37896b4bb898ce2c93327}{changes}
in the official AVA evaluation code. The evaluated models are unchanged.

\begin{table}[h!]
\centering
\vspace{-5pt}
\hspace*{-7pt}
\tablestyle{2.0pt}{1.04}
\resizebox{1.02\linewidth}{!}{
\begin{tabular}{l|c|cc|c}
{model} & pre-train & {\footnotesize center} & {\footnotesize full} & version \\ 
\shline \hline
MViT-L{\scriptsize\textuparrow312, 40\x3}~\cite{mvitv2}, \textbf{MaskFeat} & K400 & 36.3 & 37.5 & v1\\ 
MViT-L{\scriptsize\textuparrow312, 40\x3}~\cite{mvitv2}, \textbf{MaskFeat} & K600 & 37.8 & 38.8 & v1\\ 
\hline
MViT-L{\scriptsize\textuparrow312, 40\x3}~\cite{mvitv2}, \textbf{MaskFeat} & K400 & 37.3 & 38.5 & v2 \\ 
MViT-L{\scriptsize\textuparrow312, 40\x3}~\cite{mvitv2}, \textbf{MaskFeat} & K600 & 38.8 & 39.8 & v2 \\ 
\end{tabular}
}
\vspace{-5pt}
\caption{\textbf{Updated AVA results} in \cref{tab:ava} according to changes in the official AVA evaluation code. The models are unchanged.}
\label{tab:ava-v2}
\end{table}

{\small
\bibliographystyle{ieee_fullname}
\bibliography{egbib}

\begin{thebibliography}{100}\itemsep=-1pt

\bibitem{vivit}
Anurag Arnab, Mostafa Dehghani, Georg Heigold, Chen Sun, Mario Lu{\v{c}}i{\'c},
  and Cordelia Schmid.
\newblock {ViViT}: A video vision transformer.
\newblock In {\em ICCV}, 2021.

\bibitem{beit}
Hangbo Bao, Li Dong, and Furu Wei.
\newblock {BEiT}: {BERT} pre-training of image transformers.
\newblock {\em arXiv preprint arXiv:2106.08254}, 2021.

\bibitem{Bau_2017_CVPR}
David Bau, Bolei Zhou, Aditya Khosla, Aude Oliva, and Antonio Torralba.
\newblock Network dissection: Quantifying interpretability of deep visual
  representations.
\newblock In {\em CVPR}, 2017.

\bibitem{timesformer}
Gedas Bertasius, Heng Wang, and Lorenzo Torresani.
\newblock Is space-time attention all you need for video understanding?
\newblock In {\em ICML}, 2021.

\bibitem{nfnet}
Andrew Brock, Soham De, Samuel~L Smith, and Karen Simonyan.
\newblock High-performance large-scale image recognition without normalization.
\newblock In {\em ICML}, 2021.

\bibitem{gpt-3}
Tom~B. Brown, Benjamin Mann, Nick Ryder, Melanie Subbiah, Jared Kaplan,
  Prafulla Dhariwal, Arvind Neelakantan, Pranav Shyam, Girish Sastry, Amanda
  Askell, Sandhini Agarwal, Ariel Herbert-Voss, Gretchen Krueger, Tom Henighan,
  Rewon Child, Aditya Ramesh, Daniel~M. Ziegler, Jeffrey Wu, Clemens Winter,
  Christopher Hesse, Mark Chen, Eric Sigler, Mateusz Litwin, Scott Gray,
  Benjamin Chess, Jack Clark, Christopher Berner, Sam McCandlish, Alec Radford,
  Ilya Sutskever, and Dario Amodei.
\newblock Language models are few-shot learners.
\newblock In {\em NeurIPS}, 2020.

\bibitem{deepclustering}
Mathilde Caron, Piotr Bojanowski, Armand Joulin, and Matthijs Douze.
\newblock Deep clustering for unsupervised learning of visual features.
\newblock In {\em ECCV}, 2018.

\bibitem{swav}
Mathilde Caron, Ishan Misra, Julien Mairal, Priya Goyal, Piotr Bojanowski, and
  Armand Joulin.
\newblock Unsupervised learning of visual features by contrasting cluster
  assignments.
\newblock In {\em NeurIPS}, 2020.

\bibitem{dino}
Mathilde Caron, Hugo Touvron, Ishan Misra, Herv\'e J\'egou, Julien Mairal,
  Piotr Bojanowski, and Armand Joulin.
\newblock Emerging properties in self-supervised vision transformers.
\newblock In {\em ICCV}, 2021.

\bibitem{k600}
Joao Carreira, Eric Noland, Andras Banki-Horvath, Chloe Hillier, and Andrew
  Zisserman.
\newblock A short note about kinetics-600.
\newblock {\em arXiv preprint arXiv:1808.01340}, 2018.

\bibitem{k700}
Joao Carreira, Eric Noland, Chloe Hillier, and Andrew Zisserman.
\newblock A short note on the kinetics-700 human action dataset.
\newblock {\em arXiv preprint arXiv:1907.06987}, 2019.

\bibitem{two-stream}
Joao Carreira and Andrew Zisserman.
\newblock Quo vadis, action recognition? a new model and the kinetics dataset.
\newblock In {\em CVPR}, 2017.

\bibitem{chatfield2014return}
Ken Chatfield, Karen Simonyan, Andrea Vedaldi, and Andrew Zisserman.
\newblock Return of the devil in the details: Delving deep into convolutional
  nets.
\newblock In {\em BMVC}, 2014.

\bibitem{igpt}
Mark Chen, Alec Radford, Rewon Child, Jeff Wu, Heewoo Jun, Prafulla Dhariwal,
  David Luan, and Ilya Sutskever.
\newblock Generative pretraining from pixels.
\newblock In {\em ICML}, 2020.

\bibitem{simclr}
Ting Chen, Simon Kornblith, Mohammad Norouzi, and Geoffrey Hinton.
\newblock A simple framework for contrastive learning of visual
  representations.
\newblock In {\em ICML}, 2020.

\bibitem{mocov2}
Xinlei Chen, Haoqi Fan, Ross Girshick, and Kaiming He.
\newblock Improved baselines with momentum contrastive learning.
\newblock {\em arXiv preprint arXiv:2003.04297}, 2020.

\bibitem{simsiam}
Xinlei Chen and Kaiming He.
\newblock Exploring simple siamese representation learning.
\newblock In {\em CVPR}, 2021.

\bibitem{mocov3}
Xinlei Chen, Saining Xie, and Kaiming He.
\newblock An empirical study of training self-supervised vision transformers.
\newblock In {\em ICCV}, 2021.

\bibitem{uniter}
Yen-Chun Chen, Linjie Li, Licheng Yu, Ahmed~El Kholy, Faisal Ahmed, Zhe Gan, Yu
  Cheng, and Jingjing Liu.
\newblock Uniter: Universal image-text representation learning.
\newblock In {\em ECCV}, 2020.

\bibitem{electra}
Kevin Clark, Minh-Thang Luong, Quoc~V Le, and Christopher~D Manning.
\newblock {ELECTRA}: Pre-training text encoders as discriminators rather than
  generators.
\newblock In {\em ICLR}, 2020.

\bibitem{randaug}
Ekin~D Cubuk, Barret Zoph, Jonathon Shlens, and Quoc~V Le.
\newblock {RandAugment}: Practical automated data augmentation with a reduced
  search space.
\newblock In {\em CVPR}, 2020.

\bibitem{hog}
Navneet Dalal and Bill Triggs.
\newblock Histograms of oriented gradients for human detection.
\newblock In {\em CVPR}, 2005.

\bibitem{hogvideo}
Navneet Dalal, Bill Triggs, and Cordelia Schmid.
\newblock Human detection using oriented histograms of flow and appearance.
\newblock In {\em ECCV}, 2006.

\bibitem{imagenet}
Jia Deng, Wei Dong, Richard Socher, Li-Jia Li, Kai Li, and Li Fei-Fei.
\newblock {ImageNet}: A large-scale hierarchical image database.
\newblock In {\em CVPR}, 2009.

\bibitem{bert}
Jacob Devlin, Ming-Wei Chang, Kenton Lee, and Kristina Toutanova.
\newblock {BERT}: Pre-training of deep bidirectional transformers for language
  understanding.
\newblock In {\em NAACL}, 2019.

\bibitem{context-prediction}
Carl Doersch, Abhinav Gupta, and Alexei~A Efros.
\newblock Unsupervised visual representation learning by context prediction.
\newblock In {\em ICCV}, 2015.

\bibitem{vit}
Alexey Dosovitskiy, Lucas Beyer, Alexander Kolesnikov, Dirk Weissenborn,
  Xiaohua Zhai, Thomas Unterthiner, Mostafa Dehghani, Matthias Minderer, Georg
  Heigold, Sylvain Gelly, Jakob Uszkoreit, and Neil Houlsby.
\newblock An image is worth 16x16 words: Transformers for image recognition at
  scale.
\newblock In {\em ICLR}, 2021.

\bibitem{dosovitskiy2015discriminative}
Alexey Dosovitskiy, Philipp Fischer, Jost~Tobias Springenberg, Martin
  Riedmiller, and Thomas Brox.
\newblock Discriminative unsupervised feature learning with exemplar
  convolutional neural networks.
\newblock {\em TPAMI}, 2015.

\bibitem{fan2020pyslowfast}
Haoqi Fan, Yanghao Li, Bo Xiong, Wan-Yen Lo, and Christoph Feichtenhofer.
\newblock {PySlowFast}.
\newblock \url{https://github.com/facebookresearch/slowfast}, 2020.

\bibitem{fan2021pytorchvideo}
Haoqi Fan, Tullie Murrell, Heng Wang, Kalyan~Vasudev Alwala, Yanghao Li, Yilei
  Li, Bo Xiong, Nikhila Ravi, Meng Li, Haichuan Yang, Jitendra Malik, Ross
  Girshick, Matt Feiszli, Aaron Adcock, Wan-Yen Lo, and Christoph
  Feichtenhofer.
\newblock {PyTorchVideo}: A deep learning library for video understanding.
\newblock In {\em ACM MM}, 2021.
\newblock \url{https://pytorchvideo.org/}.

\bibitem{mvit}
Haoqi Fan, Bo Xiong, Karttikeya Mangalam, Yanghao Li, Zhicheng Yan, Jitendra
  Malik, and Christoph Feichtenhofer.
\newblock Multiscale vision transformers.
\newblock In {\em ICCV}, 2021.

\bibitem{x3d}
Christoph Feichtenhofer.
\newblock {X3D}: Expanding architectures for efficient video recognition.
\newblock In {\em CVPR}, 2020.

\bibitem{slowfast}
Christoph Feichtenhofer, Haoqi Fan, Jitendra Malik, and Kaiming He.
\newblock Slowfast networks for video recognition.
\newblock In {\em ICCV}, 2019.

\bibitem{videomoco}
Christoph Feichtenhofer, Haoqi Fan, Bo Xiong, Ross Girshick, and Kaiming He.
\newblock A large-scale study on unsupervised spatiotemporal representation
  learning.
\newblock In {\em CVPR}, 2021.

\bibitem{fernando2017self}
Basura Fernando, Hakan Bilen, Efstratios Gavves, and Stephen Gould.
\newblock Self-supervised video representation learning with odd-one-out
  networks.
\newblock In {\em CVPR}, 2017.

\bibitem{rotation}
Spyros Gidaris, Praveer Singh, and Nikos Komodakis.
\newblock Unsupervised representation learning by predicting image rotations.
\newblock In {\em ICLR}, 2018.

\bibitem{fast-rcnn}
Ross Girshick.
\newblock Fast {R-CNN}.
\newblock In {\em ICCV}, 2015.

\bibitem{goroshin2015unsupervised}
Ross Goroshin, Joan Bruna, Jonathan Tompson, David Eigen, and Yann LeCun.
\newblock Unsupervised learning of spatiotemporally coherent metrics.
\newblock In {\em ICCV}, 2015.

\bibitem{1hour}
Priya Goyal, Piotr Doll{\'a}r, Ross Girshick, Pieter Noordhuis, Lukasz
  Wesolowski, Aapo Kyrola, Andrew Tulloch, Yangqing Jia, and Kaiming He.
\newblock Accurate, large minibatch {SGD}: Training imagenet in 1 hour.
\newblock {\em arXiv preprint arXiv:1706.02677}, 2017.

\bibitem{ssv2}
Raghav Goyal, Samira Ebrahimi~Kahou, Vincent Michalski, Joanna Materzynska,
  Susanne Westphal, Heuna Kim, Valentin Haenel, Ingo Fruend, Peter Yianilos,
  Moritz Mueller-Freitag, et~al.
\newblock The ``something something'' video database for learning and
  evaluating visual common sense.
\newblock In {\em ICCV}, 2017.

\bibitem{byol}
Jean-Bastien Grill, Florian Strub, Florent Altch{\'e}, Corentin Tallec,
  Pierre~H. Richemond, Elena Buchatskaya, Carl Doersch, Bernardo~Avila Pires,
  Zhaohan~Daniel Guo, Mohammad~Gheshlaghi Azar, Bilal Piot, Koray Kavukcuoglu,
  Rémi Munos, and Michal Valko.
\newblock Bootstrap your own latent: A new approach to self-supervised
  learning.
\newblock In {\em NeruIPS}, 2020.

\bibitem{ava}
Chunhui Gu, Chen Sun, Sudheendra Vijayanarasimhan, Caroline Pantofaru, David~A.
  Ross, George Toderici, Yeqing Li, Susanna Ricco, Rahul Sukthankar, Cordelia
  Schmid, and Jitendra Malik.
\newblock {AVA}: A video dataset of spatio-temporally localized atomic visual
  actions.
\newblock In {\em CVPR}, 2018.

\bibitem{mae}
Kaiming He, Xinlei Chen, Saining Xie, Yanghao Li, Piotr Doll{\'a}r, and Ross
  Girshick.
\newblock Masked autoencoders are scalable vision learners.
\newblock {\em arXiv preprint arXiv:2111.06377}, 2021.

\bibitem{moco}
Kaiming He, Haoqi Fan, Yuxin Wu, Saining Xie, and Ross Girshick.
\newblock Momentum contrast for unsupervised visual representation learning.
\newblock In {\em CVPR}, 2020.

\bibitem{mask-rcnn}
Kaiming He, Georgia Gkioxari, Piotr Doll{\'a}r, and Ross Girshick.
\newblock Mask {R-CNN}.
\newblock In {\em ICCV}, 2017.

\bibitem{resnet}
Kaiming He, Xiangyu Zhang, Shaoqing Ren, and Jian Sun.
\newblock Deep residual learning for image recognition.
\newblock In {\em CVPR}, 2016.

\bibitem{orvit}
Roei Herzig, Elad Ben-Avraham, Karttikeya Mangalam, Amir Bar, Gal Chechik, Anna
  Rohrbach, Trevor Darrell, and Amir Globerson.
\newblock Object-region video transformers.
\newblock {\em arXiv preprint arXiv:2110.06915}, 2021.

\bibitem{distillation}
Geoffrey Hinton, Oriol Vinyals, and Jeff Dean.
\newblock Distilling the knowledge in a neural network.
\newblock In {\em NeurIPS}, 2015.

\bibitem{hinton2015distilling}
Geoffrey Hinton, Oriol Vinyals, and Jeff Dean.
\newblock Distilling the knowledge in a neural network.
\newblock {\em arXiv preprint arXiv:1503.02531}, 2015.

\bibitem{tokenlabeling}
Zihang Jiang, Qibin Hou, Li Yuan, Daquan Zhou, Yujun Shi, Xiaojie Jin, Anran
  Wang, and Jiashi Feng.
\newblock All tokens matter: Token labeling for training better vision
  transformers.
\newblock In {\em NeurIPS}, 2021.

\bibitem{k400}
Will Kay, Joao Carreira, Karen Simonyan, Brian Zhang, Chloe Hillier, Sudheendra
  Vijayanarasimhan, Fabio Viola, Tim Green, Trevor Back, Paul Natsev, et~al.
\newblock The kinetics human action video dataset.
\newblock {\em arXiv preprint arXiv:1705.06950}, 2017.

\bibitem{vilt}
Wonjae Kim, Bokyung Son, and Ildoo Kim.
\newblock {ViLT}: Vision-and-language transformer without convolution or region
  supervision.
\newblock In {\em ICML}, 2021.

\bibitem{movinet}
Dan Kondratyuk, Liangzhe Yuan, Yandong Li, Li Zhang, Mingxing Tan, Matthew
  Brown, and Boqing Gong.
\newblock {MoviNets}: Mobile video networks for efficient video recognition.
\newblock In {\em CVPR}, 2021.

\bibitem{droppath}
Gustav Larsson, Michael Maire, and Gregory Shakhnarovich.
\newblock {FractalNet}: Ultra-deep neural networks without residuals.
\newblock In {\em ICLR}, 2017.

\bibitem{srgan}
Christian Ledig, Lucas Theis, Ferenc Husz{\'a}r, Jose Caballero, Andrew
  Cunningham, Alejandro Acosta, Andrew Aitken, Alykhan Tejani, Johannes Totz,
  Zehan Wang, et~al.
\newblock Photo-realistic single image super-resolution using a generative
  adversarial network.
\newblock In {\em CVPR}, 2017.

\bibitem{mvitv2}
Yanghao Li, Chao-Yuan Wu, Haoqi Fan, Karttikeya Mangalam, Bo Xiong, Jitendra
  Malik, and Christoph Feichtenhofer.
\newblock Improved multiscale vision transformers for classification and
  detection.
\newblock {\em arXiv preprint arXiv:2112.01526}, 2021.

\bibitem{tsm}
Ji Lin, Chuang Gan, and Song Han.
\newblock {TSM}: Temporal shift module for efficient video understanding.
\newblock In {\em ICCV}, 2019.

\bibitem{swinv2}
Ze Liu, Han Hu, Yutong Lin, Zhuliang Yao, Zhenda Xie, Yixuan Wei, Jia Ning, Yue
  Cao, Zheng Zhang, Li Dong, Furu Wei, and Baining Guo.
\newblock Swin transformer v2: Scaling up capacity and resolution.
\newblock {\em arXiv preprint arXiv:2111.09883}, 2021.

\bibitem{video-swin}
Ze Liu, Jia Ning, Yue Cao, Yixuan Wei, Zheng Zhang, Stephen Lin, and Han Hu.
\newblock Video swin transformer.
\newblock {\em arXiv preprint arXiv:2106.13230}, 2021.

\bibitem{sgdr}
Ilya Loshchilov and Frank Hutter.
\newblock {SGDR}: Stochastic gradient descent with warm restarts.
\newblock In {\em ICLR}, 2017.

\bibitem{adamw}
Ilya Loshchilov and Frank Hutter.
\newblock Decoupled weight decay regularization.
\newblock In {\em ICLR}, 2019.

\bibitem{sift}
David~G Lowe.
\newblock Object recognition from local scale-invariant features.
\newblock In {\em ICCV}, 1999.

\bibitem{vilbert}
Jiasen Lu, Dhruv Batra, Devi Parikh, and Stefan Lee.
\newblock {ViLBERT}: Pretraining task-agnostic visiolinguistic representations
  for vision-and-language tasks.
\newblock In {\em NeurIPS}, 2019.

\bibitem{misra2016shuffle}
Ishan Misra, C~Lawrence Zitnick, and Martial Hebert.
\newblock Shuffle and learn: unsupervised learning using temporal order
  verification.
\newblock In {\em ECCV}, 2016.

\bibitem{jigsaw}
Mehdi Noroozi and Paolo Favaro.
\newblock Unsupervised learning of visual representations by solving jigsaw
  puzzles.
\newblock In {\em ECCV}, 2016.

\bibitem{acar}
Junting Pan, Siyu Chen, Mike~Zheng Shou, Yu Liu, Jing Shao, and Hongsheng Li.
\newblock Actor-context-actor relation network for spatio-temporal action
  localization.
\newblock In {\em CVPR}, 2021.

\bibitem{pytorch}
Adam Paszke, Sam Gross, Francisco Massa, Adam Lerer, James Bradbury, Gregory
  Chanan, Trevor Killeen, Zeming Lin, Natalia Gimelshein, Luca Antiga, Alban
  Desmaison, Andreas Kopf, Edward Yang, Zachary DeVito, Martin Raison, Alykhan
  Tejani, Sasank Chilamkurthy, Benoit Steiner, Lu Fang, Junjie Bai, and Soumith
  Chintala.
\newblock {PyTorch}: An imperative style, high-performance deep learning
  library.
\newblock In {\em NeurIPS}, 2019.

\bibitem{pathak2017learning}
Deepak Pathak, Ross Girshick, Piotr Doll{\'a}r, Trevor Darrell, and Bharath
  Hariharan.
\newblock Learning features by watching objects move.
\newblock In {\em CVPR}, 2017.

\bibitem{inpainting}
Deepak Pathak, Philipp Krahenbuhl, Jeff Donahue, Trevor Darrell, and Alexei~A
  Efros.
\newblock Context encoders: Feature learning by inpainting.
\newblock In {\em CVPR}, 2016.

\bibitem{mformer}
Mandela Patrick, Dylan Campbell, Yuki~M. Asano, Ishan Misra~Florian Metze,
  Christoph Feichtenhofer, Andrea Vedaldi, and Jo{\~a}o~F. Henriques.
\newblock Keeping your eye on the ball: Trajectory attention in video
  transformers.
\newblock In {\em NeurIPS}, 2021.

\bibitem{purushwalkam2020demystifying}
Senthil Purushwalkam and Abhinav Gupta.
\newblock Demystifying contrastive self-supervised learning: Invariances,
  augmentations and dataset biases.
\newblock In {\em NeurIPS}, 2020.

\bibitem{cvrl}
Rui Qian, Tianjian Meng, Boqing Gong, Ming-Hsuan Yang, Huisheng Wang, Serge
  Belongie, and Yin Cui.
\newblock Spatiotemporal contrastive video representation learning.
\newblock In {\em CVPR}, 2021.

\bibitem{dall-e}
Aditya Ramesh, Mikhail Pavlov, Gabriel Goh, Scott Gray, Chelsea Voss, Alec
  Radford, Mark Chen, and Ilya Sutskever.
\newblock Zero-shot text-to-image generation.
\newblock In {\em ICML}, 2021.

\bibitem{faster-rcnn}
Shaoqing Ren, Kaiming He, Ross Girshick, and Jian Sun.
\newblock Faster {R-CNN}: Towards real-time object detection with region
  proposal networks.
\newblock In {\em NeurIPS}, 2015.

\bibitem{tokenlearner}
Michael~S Ryoo, AJ Piergiovanni, Anurag Arnab, Mostafa Dehghani, and Anelia
  Angelova.
\newblock Tokenlearner: What can 8 learned tokens do for images and videos?
\newblock {\em arXiv preprint arXiv:2106.11297}, 2021.

\bibitem{improved-gan}
Tim Salimans, Ian Goodfellow, Wojciech Zaremba, Vicki Cheung, Alec Radford, and
  Xi Chen.
\newblock Improved techniques for training {GANs}.
\newblock In {\em NeurIPS}, 2016.

\bibitem{relpos}
Peter Shaw, Jakob Uszkoreit, and Ashish Vaswani.
\newblock Self-attention with relative position representations.
\newblock {\em arXiv preprint arXiv:1803.02155}, 2018.

\bibitem{dropout}
Nitish Srivastava, Geoffrey Hinton, Alex Krizhevsky, Ilya Sutskever, and Ruslan
  Salakhutdinov.
\newblock Dropout: A simple way to prevent neural networks from overfitting.
\newblock {\em JMLR}, 2014.

\bibitem{vlbert}
Weijie Su, Xizhou Zhu, Yue Cao, Bin Li, Lewei Lu, Furu Wei, and Jifeng Dai.
\newblock {VL-BERT}: Pre-training of generic visual-linguistic representations.
\newblock In {\em ICLR}, 2020.

\bibitem{jft}
Chen Sun, Abhinav Shrivastava, Saurabh Singh, and Abhinav Gupta.
\newblock Revisiting unreasonable effectiveness of data in deep learning era.
\newblock In {\em ICCV}, 2017.

\bibitem{inception}
Christian Szegedy, Vincent Vanhoucke, Sergey Ioffe, Jon Shlens, and Zbigniew
  Wojna.
\newblock Rethinking the inception architecture for computer vision.
\newblock In {\em CVPR}, 2016.

\bibitem{vimpac}
Hao Tan, Jie Lei, Thomas Wolf, and Mohit Bansal.
\newblock {VIMPAC}: Video pre-training via masked token prediction and
  contrastive learning.
\newblock {\em arXiv preprint arXiv:2106.11250}, 2021.

\bibitem{cmc}
Yonglong Tian, Dilip Krishnan, and Phillip Isola.
\newblock Contrastive multiview coding.
\newblock In {\em ECCV}, 2020.

\bibitem{deit}
Hugo Touvron, Matthieu Cord, Matthijs Douze, Francisco Massa, Alexandre
  Sablayrolles, and Herve Jegou.
\newblock Training data-efficient image transformers \& distillation through
  attention.
\newblock In {\em ICML}, 2021.

\bibitem{ipcsn}
Du Tran, Heng Wang, Lorenzo Torresani, and Matt Feiszli.
\newblock Video classification with channel-separated convolutional networks.
\newblock In {\em ICCV}, 2019.

\bibitem{colorhog}
Koen Van De~Sande, Theo Gevers, and Cees Snoek.
\newblock Evaluating color descriptors for object and scene recognition.
\newblock {\em TPAMI}, 2009.

\bibitem{vqvae}
Aaron van~den Oord, Oriol Vinyals, and Koray Kavukcuoglu.
\newblock Neural discrete representation learning.
\newblock In {\em NeurIPS}, 2017.

\bibitem{transformer}
Ashish Vaswani, Noam Shazeer, Niki Parmar, Jakob Uszkoreit, Llion Jones,
  Aidan~N Gomez, {\L}ukasz Kaiser, and Illia Polosukhin.
\newblock Attention is all you need.
\newblock In {\em NeurIPS}, 2017.

\bibitem{denoisingae}
Pascal Vincent, Hugo Larochelle, Isabelle Lajoie, Yoshua Bengio, Pierre-Antoine
  Manzagol, and L{\'e}on Bottou.
\newblock Stacked denoising autoencoders: Learning useful representations in a
  deep network with a local denoising criterion.
\newblock {\em JMLR}, 2010.

\bibitem{wang2017transitive}
Xiaolong Wang, Kaiming He, and Abhinav Gupta.
\newblock Transitive invariance for self-supervised visual representation
  learning.
\newblock In {\em ICCV}, 2017.

\bibitem{jigsaw2}
Chen Wei, Lingxi Xie, Xutong Ren, Yingda Xia, Chi Su, Jiaying Liu, Qi Tian, and
  Alan~L Yuille.
\newblock Iterative reorganization with weak spatial constraints: Solving
  arbitrary jigsaw puzzles for unsupervised representation learning.
\newblock In {\em CVPR}, 2019.

\bibitem{object-transformer}
Chao-Yuan Wu and Philipp Krahenbuhl.
\newblock Towards long-form video understanding.
\newblock In {\em CVPR}, 2021.

\bibitem{instdist}
Zhirong Wu, Yuanjun Xiong, X~Yu Stella, and Dahua Lin.
\newblock Unsupervised feature learning via non-parametric instance
  discrimination.
\newblock In {\em CVPR}, 2018.

\bibitem{xiao2020should}
Tete Xiao, Xiaolong Wang, Alexei~A Efros, and Trevor Darrell.
\newblock What should not be contrastive in contrastive learning.
\newblock In {\em ICLR}, 2020.

\bibitem{florence}
Lu Yuan, Dongdong Chen, Yi-Ling Chen, Noel Codella, Xiyang Dai, Jianfeng Gao,
  Houdong Hu, Xuedong Huang, Boxin Li, Chunyuan Li, Ce Liu, Mengchen Liu,
  Zicheng Liu, Yumao Lu, Yu Shi, Lijuan Wang, Jianfeng Wang, Bin Xiao, Zhen
  Xiao, Jianwei Yang, Michael Zeng, Luowei Zhou, and Pengchuan Zhang.
\newblock Florence: A new foundation model for computer vision.
\newblock {\em arXiv preprint arXiv:2111.11432}, 2021.

\bibitem{cutmix}
Sangdoo Yun, Dongyoon Han, Seong~Joon Oh, Sanghyuk Chun, Junsuk Choe, and
  Youngjoon Yoo.
\newblock Cutmix: Regularization strategy to train strong classifiers with
  localizable features.
\newblock In {\em ICCV}, 2019.

\bibitem{mixup}
Hongyi Zhang, Moustapha Cisse, Yann~N Dauphin, and David Lopez-Paz.
\newblock mixup: Beyond empirical risk minimization.
\newblock In {\em ICLR}, 2018.

\bibitem{colorization}
Richard Zhang, Phillip Isola, and Alexei~A Efros.
\newblock Colorful image colorization.
\newblock In {\em ECCV}, 2016.

\bibitem{good-transfer}
Nanxuan Zhao, Zhirong Wu, Rynson W.~H. Lau, and Stephen Lin.
\newblock What makes instance discrimination good for transfer learning?
\newblock In {\em ICLR}, 2021.

\bibitem{zhou2014object}
Bolei Zhou, Aditya Khosla, Agata Lapedriza, Aude Oliva, and Antonio Torralba.
\newblock Object detectors emerge in deep scene cnns.
\newblock In {\em ICLR}, 2015.

\end{thebibliography}
}

\end{document}